\documentclass{ieeeaccess}
\usepackage{cite}
\usepackage{amsmath,amssymb,amsfonts}
\usepackage{textcomp}
\usepackage{algorithm} 
\usepackage{algpseudocode} 
\usepackage{booktabs}
 \usepackage{multirow}
 \usepackage{graphicx}
\usepackage{url}

\usepackage{listings}
\usepackage{lipsum}

\definecolor{codegreen}{rgb}{0,0.6,0}
\definecolor{codegray}{rgb}{0.5,0.5,0.5}
\definecolor{codepurple}{rgb}{0.58,0,0.82}
\definecolor{backcolour}{rgb}{1,1,1}

\lstdefinestyle{mystyle}{
  backgroundcolor=\color{backcolour},   commentstyle=\color{codegreen},
  keywordstyle=\color{magenta},
  numberstyle=\tiny\color{codegray},
  stringstyle=\color{codepurple},
  basicstyle=\ttfamily\footnotesize,
  breakatwhitespace=false,         
  breaklines=true,                 
  captionpos=b,                    
  keepspaces=true,                 
  numbers=left,                    
  numbersep=5pt,                  
  showspaces=false,                
  showstringspaces=false,
  showtabs=false,                  
  tabsize=2
}

\algnewcommand\algorithmicforeach{\textbf{for each}}
\algdef{S}[FOR]{ForEach}[1]{\algorithmicforeach\ #1\ \algorithmicdo}

\def\BibTeX{{\rm B\kern-.05em{\sc i\kern-.025em b}\kern-.08em
    T\kern-.1667em\lower.7ex\hbox{E}\kern-.125emX}}

\definecolor{red}{cmyk}{1,0,0,1}
\definecolor{blue}{cmyk}{0,0,1,1}
\definecolor{orange}{cmyk}{0,0.5,1,0}

\newcommand{\method}{FuSS}

\usepackage{xparse}

\makeatletter
\NewDocumentCommand{\LeftComment}{s m}{%
  \Statex \IfBooleanF{#1}{\ \ \ \ \ \ \hspace*{\ALG@thistlm}}\(\triangleright\) #2}
\makeatother

\begin{document}
\doi{}

\title{\method: Fusing Superpixels for Improved Segmentation Consistency}
\author{\uppercase{Ian Nunes}\authorrefmark{1}, 
\uppercase{Matheus B. Pereira}\authorrefmark{2}, \uppercase{Hugo Oliveira}\authorrefmark{3}, \uppercase{Jefersson A. dos Santos}\authorrefmark{2} and \uppercase{Marcus Poggi}\authorrefmark{1}
}
\address[1]{Department of Informatics, at Pontifical Catholic University of Rio de Janeiro, Rio de Janeiro, Brazil (e-mail: inunes@inf.puc-rio.br)}
\address[2]{Department of Computer Science (DCC) in Universidade Federal de Minas Gerais (UFMG), Belo Horizonte, Brazil (e-mail: matheuspereira@dcc.ufmg.br)}
\address[3]{Institute of Mathematics and Statistics (IME), at University of S\~{a}o Paulo (USP), S\~{a}o Paulo, Brazil (e-mail: oliveirahugo@ime.usp.br)}
\tfootnote{This research was partially financed by the Coordenação de Aperfeiçoamento de Pessoal de Nível Superior (CAPES), Fundação de Amparo à Pesquisa do Estado de Minas Gerais (FAPEMIG), Fundação de Amparo à Pesquisa do Estado de São Paulo (FAPESP -- grant \#2020/06744-5), Agence Nationale de la Recherche (ANR) and the Serrapilheira Institute (grant R-2011-37776). The authors would also like to thank NVIDIA for the donation of the GPUs used in this research.}

\markboth
{Nunes \headeretal: FuSS: Fusing Superpixels for Improved
Segmentation Consistency}
{Nunes \headeretal: FuSS: Fusing Superpixels for Improved
Segmentation Consistency}

\corresp{Corresponding author: Ian Nunes (e-mail: inunes@inf.puc-rio.br).}

\begin{abstract}
In this work, we propose two different approaches to improve the semantic consistency of Open Set Semantic Segmentation. First, we propose a method called OpenGMM that extends the OpenPCS framework using a Gaussian Mixture of Models to model the distribution of pixels for each class in a multimodal manner. The second approach is a post-processing which uses superpixels to enforce highly homogeneous regions to behave equally, rectifying erroneous classified pixels within these regions, we also proposed a novel superpixel method called FuSS. All tests were performed on ISPRS Vaihingen and Potsdam datasets, and both methods were capable to improve quantitative and qualitative results for both datasets. Besides that, the post-process with FuSS achieved state-of-the-art results for both datasets. The official implementation is available at: \url{https://github.com/iannunes/FuSS}.
\end{abstract}


\titlepgskip=-15pt

\maketitle

\section{Introduction}
\label{sec:introduction}


Remote sensing acquisition technologies have been in constant development since the 1960s, providing sensors with a myriad of new electromagnetic spectral encoding capabilities and leading to a continuous increase in the volume of daily collected data. The development of these technologies, together with an increased capacity to produce relevant information for a wide range of applications, turned the automatic analysis of images into one of the most actively researched fields within the remote sensing community~\cite{paoletti2018capsule}.

Since the Gestalt movement, it is known that image segmentation and clustering plays an important role in human perception \cite{ellis2013source}. Many different 
applications can benefit from semantic segmentation of remote sensing images, such as: urban planning/mapping and change detection \cite{konstantinidis2017building}, population estimation \cite{almeida2018dealing}, real-estate management \cite{li2018building}.

\begin{figure*}[!th]
    \centering
    \includegraphics[width=0.7\textwidth]{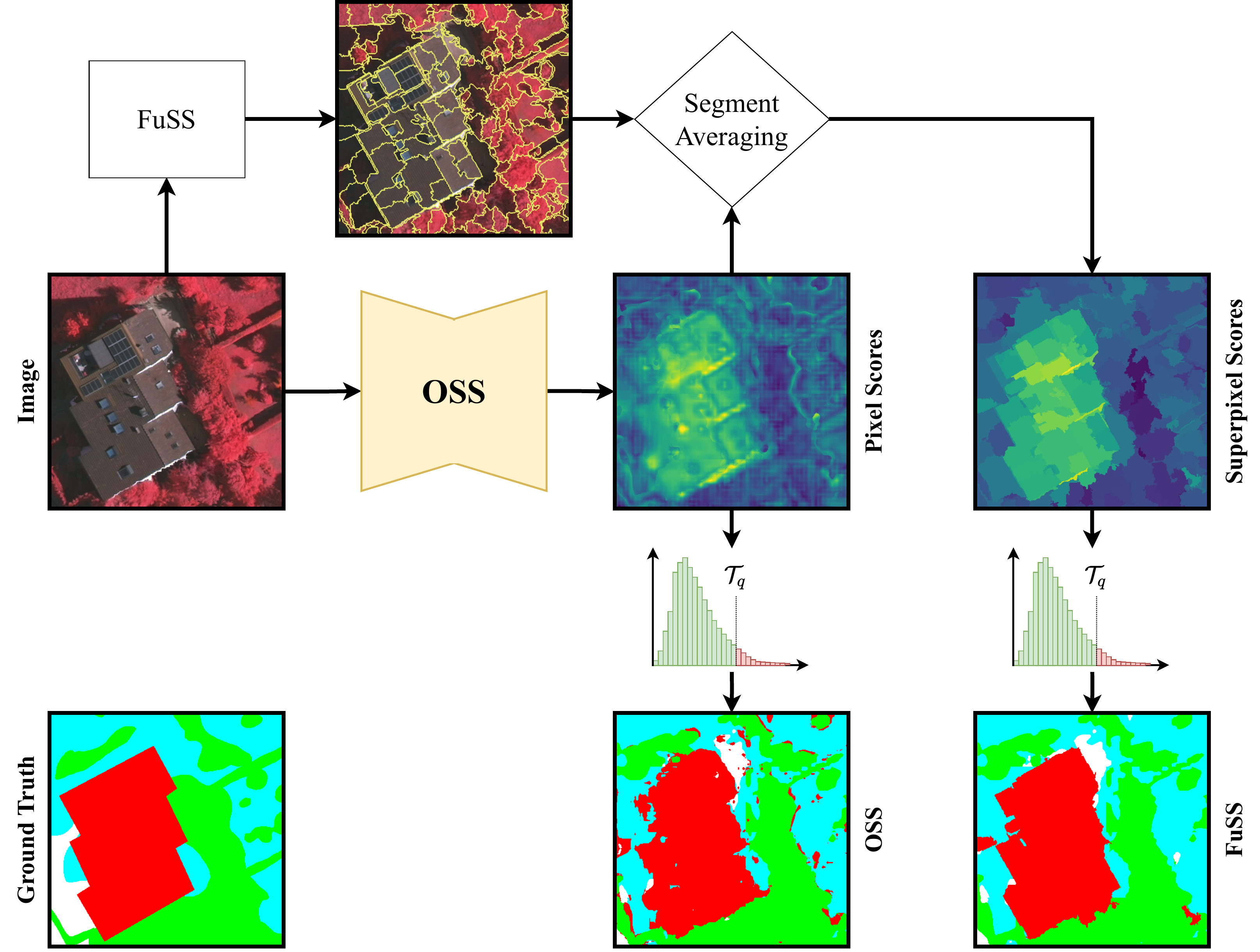}
    \caption{Post-processing procedure for OSS proposed in this work. First the input image is processed by the chosen open set segmentation method. Afterwards, the likelihood scores generated by the OSS method (e.g. reconstruction errors, likelihood scores, etc) are processed using superpixel segmentation and the mean score of each segment score is set to the whole area. The final step includes thresholding the predictions according to some cutoff -- usually set according to quantiles on the scores -- to identify unknown superpixels.}
    \label{fig:graphical_abstract}
\end{figure*}

Traditional (closed set) semantic segmentation (SSeg) for remote sensing imagery has become a complex, time-consuming, and well-studied task. Many different methods have been developed to solve this problem: FCN \cite{long2015fully}, U-Net \cite{ronneberger2015u}, SegNet \cite{badrinarayanan2017segnet}, among others. In recent years, many models have significantly improved the quality of semantic segmentation \cite{minaee2021image, ulku2022survey}.

To enable the use of SSeg models in practical situations, it is crucial to improve the semantic and spatial consistency of the resultant segmentation. 
According to Sekkal \textit{et al.} \cite{sekkal2012fast}, it is important for the regions to remain coherent with the original content and the better detection of contours leads to an efficient pseudo-semantic representation.

In closed-set SSeg, both the training and test data share the same label and feature spaces. However, in more realistic scenarios unseen classes may show up on the deploy phase of the model. This is the case for most real-world applications, such as autonomous vehicles, medical diagnosis or treatment, and remote sensing tasks. The existence of unknown classes undermines the robustness of the existing closed set methods, as stated by Geng \textit{et al.} \cite{geng2020recent}.

For the last couple of years, methods that extend traditional closed set SSeg were proposed to automatically recognize samples from unseen classes. This new task, named Open Set Segmentation (OSS), 
must be able to correctly segment pixels of classes available during training (Known Known Classes -- KKCs), while also being able to recognize unknown pixels that come from classes that were not present during training (Unknown Unknown Classes -- UUCs) \cite{oliveira2021fully}.

The goal of the current paper is to improve the semantic consistency of previously obtained results OSS algorithms for remote sensing images. 
In order to improve segmentation consistency, we introduce in this work a novel method called OpenGMM based on previously proposed frameworks for OSR \cite{oliveira2021fully,vendramini2021opening}. Oliveira \textit{et al.} \cite{oliveira2021fully} employed Principal Component Analysis (PCA) in the OpenPCS method as a generative model, assuming it would be sufficient for the representation of the data. As data in the real world can rarely be represented by unimodal distributions, our proposed approach adopts a Gaussian Mixture Model (GMM) \cite{rasmussen2003gaussian, bishop2006pattern} instead of PCA, allowing the use of multimodal distributions for modeling KKCs with the objective of improving the out-of-distribution (OOD) identification.

The lack of semantic consistency produced by OpenPCS, specially in objects borders, can be seen in Figure~\ref{fig:graphical_abstract}. Threshold methods like OpenPCS presents difficulties in correctly delimiting the edges between objects making these kind of OSS approaches not likely to be used in real world scenarios. To address this issue and improve the final results we studied ways to add a post-process using Superpixel Segmentation (SPS) with qualitative results shown in Figure~\ref{fig:graphical_abstract}, the post processed OSS is closer to the ground truth and many generated artifacts were removed.





Superpixels are a oversegmentation of an image and played an important role in the process of traditional segmentation, and acoording to Lin \textit{et al.} \cite{lin2021deep} are a low-level image information subdivisions. As stated by \cite{felzenszwalb2004efficient, verelst2019generating, wang2019content} the use of superpixels: allows for a richer feature representation for each segment (textural, color, scatter, gradient-based, statistic, orientation); reduces the size of the classification problem since many pixels could be represented by each cluster; even being an oversegmentation of the image the superpixel clustering produces homogeneous regions that adds semantics to each superpixel. Superpixels are a homogeneous and contiguous group of pixels in an image extracting perceptually relevant regions \cite{felzenszwalb2004efficient}. The segmentation generated with superpixels reduces the input size, while preserving the semantic content needed to address intended tasks \cite{verelst2019generating}. 

Our framework employs a novel superpixel merge procedure using Malahanobis distance \cite{mahalanobis1936generalized}, called \textbf{Fu}sing \textbf{S}uperpixels for \textbf{S}egmentation (\method). A graphical illustration of our pipeline's inner workings can be seen in Figure~\ref{fig:graphical_abstract}. We tested our approach on OSS using three distinct algorithms: OpenPCS \cite{oliveira2021fully}, CoReSeg \cite{nunes2022conditional} and the newly proposed OpenGMM. 

Our pipeline seeks to improve the semantic consistency of the results in order to make OSS models more suitable for practical use. Our contributions can be summarized as: 
\begin{itemize}
    \item The proposal of OpenGMM, a novel method aiming to improve upon the previously proposed OSS framework by Oliveira \textit{et al.} \cite{oliveira2021fully};
    \item The proposal of a superpixel post-processing method that yielded results with superior semantic consistency and improved Receiver Operating Characteristic metrics in all tested scenarios;
    \item A novel superpixel segmentation fusion procedure using Malahanobis distance \cite{mahalanobis1936generalized};
    \item State of the Art OSS results for the Vaihingen and Potsdam datasets\footnote{\url{https://www2.isprs.org/commissions/comm2/wg4/benchmark/}\label{foot:isprs}} using our post-processing segmentation technique;
\end{itemize} 

This manucript is organized as follows: Section~\ref{sec:related} presents previous OSS strategies and related work on superpixels for semantic segmentation; Section~\ref{sec:methodology} describes the proposed methods; Section~\ref{sec:experimental_setup} introduces the experimental setup used for this work, along with the employed datasets and metrics; Section~\ref{sec:ablation} presents exploratory tests executed on Vaihingen dataset; Section~\ref{sec:results} presents the obtained results; and Section~\ref{sec:conclusion} exposes the conclusions over the proposed methods and the achieved results.

\section{Related Work}
\label{sec:related}

\subsection{Semantic Consistency in Segmentation}
\label{sec:semantic_consistency}

Semantic consistency is rarely explicitly addressed in semantic segmentation papers. In the following lines we present an overview of the few existing trends on deep semantic consistency. 

Through an end-to-end trainable network that combines 2 branches, one for edge detection and one for traditional semantic segmentation, Ji \textit{et al.} \cite{ji2020parallel} managed to improve the performance and the spatial consistency of the resulting segmentation for PASCAL VOC 2012, PASCAL-Context and Cityscapes datasets.

PixMatch, proposed by Melas-Kyriazi \textit{et al.} \cite{melas2021pixmatch}, uses heavy augmentation and a loss term composed by the summation of two cross-entropy terms. The first loss term is standard for SSeg, and the second is calculated over a slightly perturbed image and mask. The new loss enforces the notion of smoothness in the target domain in order to enhance intra-object segmentation consistency.

Pixelwise Contrast and Consistency Learning (PiCoCo), proposed by Kang \textit{et al.} \cite{kang2021picoco}, seeks consistency in closed set semantic segmentation using a joint loss function that is summation of a supervised loss term, a contrast loss term, and a consistency loss term. The supervised is a standard semantic segmentation loss term composed by a Cross Entropy and a Dice loss term; for the contrastive loss a selection of positive and negative samples enforce the model to improve its generalization capabilities; the consistency loss term consists in a summation of a cross entropy and a dice loss of heavy augmented pairs of input and labels to enforce semantic consistency and robustness to the learning process. 

The use of a post-processing that combines unsupervised colorization and deep edge superpixels to enhance the semantic segmentation of panchromatic aerial images was proposed by Ratajczak \textit{et al.} \cite{ratajczak2020semantic}. The authors propose to assess if applying a colorization algorithm could improve the strenght of the pairwise potentials used in a conditional random field (CRF) post-processing. In this work Deep Edge Superpixels (DES) are defined using the Watershed \cite{hu2015watershed} with the intermediate activation maps obtained before each pooling layer to the output space of a Holistically-Nested Edge Detection Nework \cite{xie2015holistically}, they use the generated SPS with mean value for intensity together with CRF to improve the final semantic consistency.

Also the use of supervoxels to improve segmentation consistency was used by Zhang \textit{et al.} \cite{zhang2014discriminative}, where a 3DCNN was used to learn discriminative hierarchical features from spatio-temporal volumes.
  
Our work introduced a post-processing for OSS that uses a SPS to improve the semantic consistency of the resultant OSS, this procedure produced better results in all tested scenarios. We also proposed a novel SPS called \method{} that bennefits from the merge of different input segmentations and produces a SPS with better results in most tested scenarios when compared to the same post-processing with the single algorithm SPS.

\subsection{Open Set Semantic Segmentation}
\label{sec:oss}

As stated by Scheirer \textit{et al.} \cite{scheirer2012toward} an open set scenario happens when unknown samples can appear in the prediction phase, meaning that at training time not all possible classes are known. This definition can be applied for each pixel in an image, extending the traditional semantic segmentation to OSS.

OSS has only a handful of published works that use neural networks. The first attempts to do open set recognition from a neural network were proposed by Bendale and Boult \cite{bendale2016towards} the first natural approach was to apply a threshold on the final output probability, identifying low probabilities as unknown. The experiments showed that this strategy was weak and not enough to deal the task. The second approach was a method called OpenMax, which presented a new model layer to replace the traditional softmax. Assuming $C$ classes trained on the closed set task, OpenMax adds an Unknown output class and estimates the probability of the input images to each of the $C+1$ classes.

Based on OpenMax \cite{bendale2016towards}, OpenPixel was proposed by Silva \textit{et al.} \cite{da2020towards}. It uses a patch-wise strategy to classify the central pixel. In practice, each patch only classifies the central pixel, thus the network needs one patch per pixel to segment the whole image. OpenPixel is, therefore, extremely inefficient mainly during the test phase, as each pixel in an image generates a patch to be fed to the network, with surrounding pixels being forwarded multiple times through the backbone. The fully convolutional counterpart to OpenPixel, named OpenFCN, was proposed by Oliveira \textit{et al.} \cite{oliveira2021fully}. OpenMax-based methods proved to have their efficacy severely limited in segmentation settings, resulting in false positive OOD pixel predictions mainly in the borders of objects, where the activations of the last layers are considerably affected by the presence of surrounding objects with different classes.

Given the limitations of OpenPixel and OpenFCN caused by last layers' activations, Oliveira \textit{et al.} \cite{oliveira2021fully} proposed the use of intermediate multiscale features from the closed set FCNs paired with low-dimensional principal component scoring for OSS. The method, named Open Principal Component Scoring (OpenPCS), achieved consistently better results than the OpenMax-based approaches in multiple scenarios and unknown class configurations on the Vaihingen, Potsdam and GRSS 2018 Data Fusion Challenge \cite{GRSS} datasets. 

Chamorro \textit{et al.} \cite{martinez2021open} built upon OpenPCS, including a whitening operation to better generalize to a wide range of imbalanced classes. OpenPCS++ was tested on multi-temporal SAR images from the LEM\footnote{http://www.lvc.ele.puc-rio.br/LEM\_benchmark/} and CV\footnote{https://ieee-dataport.org/documents/campo-verde-database} datasets in the task of segmenting OOD crops among known crop species.

Cui \textit{et al.} \cite{cui2020open} proposed a nonparametric statistical OSS method that employs the Mann-Whitney U test on a closed set segmentation output to determine the existence of unknown classes in each image. Furthermore, it uses an adaptive threshold that identify which pixels are unknown.

Nunes \textit{et al.} \cite{nunes2022conditional} proposed a fully convolutional end-to-end Conditional Reconstruction Open Set Segmentation (CoReSeg) that tackles the OSS using two network branches: a traditional closed set segmentation branch and a class conditional reconstruction of the input images according to their pixelwise mask, together the branches merges into a single open set segmentation. The traditional segmentation branch produces a standard closed set segmentation and the reconstruction error from the conditional reconstruction branch identifies which pixels are OOD.

Proposed by Cen \textit{et al.} \cite{cen2021deep} an open world semantic segmentation system used prototypes for the known classes and a Deep Metric Learning Network (DML-Net) as a feature extractor. The final classifier is replaced by a representation of the Euclidean distance of the input image to all prototypes. Low probability classes are set to unknown with the use of a Euclidean distance-based probability loss. 

The Generalized Open-set Semantic Segmentation (GOSS) method was proposed by Hong \textit{et al.} \cite{hong2022goss}. The authors were inspired by the human capacity of recognizing the previously known classes, while simultaneously identifying and grouping the unknown objects into different categories, even without previous knowledge about them. The method employs two network branches trained together in parallel: the first branch performs a SSeg for KKCs and identifies unknown pixels using Deep Metric Learning (DML) \cite{cen2021deep}
the second branch is a pixel clustering one that ignores the KKCs and generates a new segmentation mask to the image. At last, the fusion phase uses the pixels defined as unknown and the clustering to identify different objects in the unknown areas.

\subsection{Superpixel Segmentation (SPS) Algorithms}
\label{sec:superpixel_segmentation}


In this work, we consider a superpixel as a group of contiguous pixels in a given image which have been grouped together according to some criterion of homogeneity. As spatiality is crucial to any SPS, neighboring superpixels should be perceptually different. Nevertheless, non-neighboring superpixels may have similar values and shapes. All pixels inside a superpixel should be represented by the mean or median value for each image band.

SPSs are an active research area, and many different methods have been proposed to generate superpixels from an image. As examples of relevant methods proposed in the last 2 decades, we can cite: Felzenszwalb \cite{felzenszwalb2004efficient}, Quickshift \cite{vedaldi2008quick}, TurboPixels \cite{levinshtein2009turbopixels}, ERS \cite{liu2011entropy}, SLIC \cite{achanta2012slic}, GSM \cite{morerio2014generative}, Eikonal-based \cite{buyssens2014eikonal}, SEEDS \cite{bergh2012seeds}, LSC \cite{li2015superpixel}, Waterpixels \cite{machairas2015waterpixels}, BASS \cite{rubio2016bass}, SAS \cite{achanta2018scale}, SH+FDAG \cite{wang2019content}, content-based \cite{zhang2020content} and SPFCM \cite{elkhateeb2021novel}. 

Among all possible choices of SPS algorithms to use in this work we choose three algorithms that have fundamentally different strategies to generate the superpixels: SLIC \cite{achanta2012slic}, Quickshift \cite{vedaldi2008quick} and Felzenszwalb \cite{felzenszwalb2004efficient}. In the following paragraphs we briefly present these three superpixel algorithms.

\subsubsection{Simple Linear Iterative Clustering}
\label{sec:slic}

The SLIC algorithm was proposed by Achanta \textit{et al.} \cite{achanta2012slic} and its goal is to group pixels into perceptually meaningful contiguous regions. 
The method adapts the K-means algorithm \cite{lloyd1982least} to generate the superpixels. 

The operation of the algorithm is simple. It starts with $n$ centers uniformly distributed in the image. Then, it adjusts the centers' positions to the local minimum of the gradient of pixel intensity as to avoid centering the superpixel in an edge. Finally, it iterates to every center, reallocating the pixels in a fixed size window to the closest center. Limiting the search area makes the algorithm faster while still producing good results. The SLIC algorithm produces areas with great homogeneity as it is built on a K-means like approach, but it may neglect the natural edges of the image as a side effect.

\subsubsection{Quickshift}
\label{sec:quickshift}

Quickshift is a fast mode seeking algorithm proposed by Vedaldi \textit{et al.} \cite{vedaldi2008quick}. In Quickshift, every pixel is started as a superpixel, then the closest ones are put together within a defined radius distance. For each superpixel, if a pixel is out of the radius limits, a new cluster is defined and populated by the pixels inside the radius of the new centroid. Quickshift is a hierarchical clustering algorithm, the size of the clusters can be derived from the generated tree of radius values. This method do not force the pixels to be close to each other spatially, what generates highly homogeneous pixels of different sizes and shapes.

\subsubsection{Felzenszwalb}
\label{sec:felzenszwalb}

Proposed by Felzenszwalb \textit{et al.} \cite{felzenszwalb2004efficient}, the algorithm is a graph based segmentation algorithm where each vertex represents a pixel and each selected edge has some dissimilarity measure as its value. Every pixel in the image is represented in the graph, but only some of the edges are selected according to a defined criterion (e.g. $K$-nearest neighbors) in order to guarantee the intended complexity for the algorithm ($O(m\ log\ n)$, where $m$ is the number of edges and $n$ the number of vertices).

There are two presented strategies to select the edges, the first takes into account the notion of grid and connects each pixel to the 8 closest ones in the grid. The second strategy maps the image into a higher level feature space and connects the $m$ closest points in this new space. For instance, a $5$-dimensional space defined by the spatial position $x$ and $y$, and the three colors of RGB.

The algorithm explores the same idea that Kruskal \textit{et al.} \cite{kruskal1956shortest} used in their classical algorithm to find the minimum spanning tree (MST) on a graph and selects the edges in a non-descending order to generate the clusters. The clusters are generated using the intuition that they are defined by a region where the values of the edges are low and the borders of the clusters have greater dissimilarities when compared to the intra-cluster values. 

By construction, the Felzenszwalb algorithm generates clusters that vary in shapes and sizes, but strongly respects the borders of the natural objects in the image.
\section{Improving OSS Semantic Consistency}
\label{sec:methodology}

In the following section we present our proposed methodology for improving the semantic consistency of OSS. Firstly, in Section~\ref{sec:opengmm} we present an extension of the OSS framework proposed by Oliveira \textit{et al.} \cite{oliveira2021fully}, replacing the unimodal dimensionality reduction of Principal Component Analysis (PCA) with a multimodal Gaussian Mixture Model (GMM) capable of opening closed set pretrained segmentation networks with a better segmentation quality. In Section~\ref{sec:oss_superpixels} we propose a superpixel post-processing for generic OSS methods capable of improving Receiver Operating Characteristic metrics and Cohen's kappa score, as well as qualitative semantic consistency. At last, in Section~\ref{sec:fuss} we introduce a novel superpixel merge method that uses the Malahanobis \cite{mahalanobis1936generalized} distance to merge neighboring superpixels enforcing a minimum pixel count for each segment.

\subsection{OpenGMM}
\label{sec:opengmm}

\begin{figure}[!ht]
    \centering
    \includegraphics[width=\columnwidth]{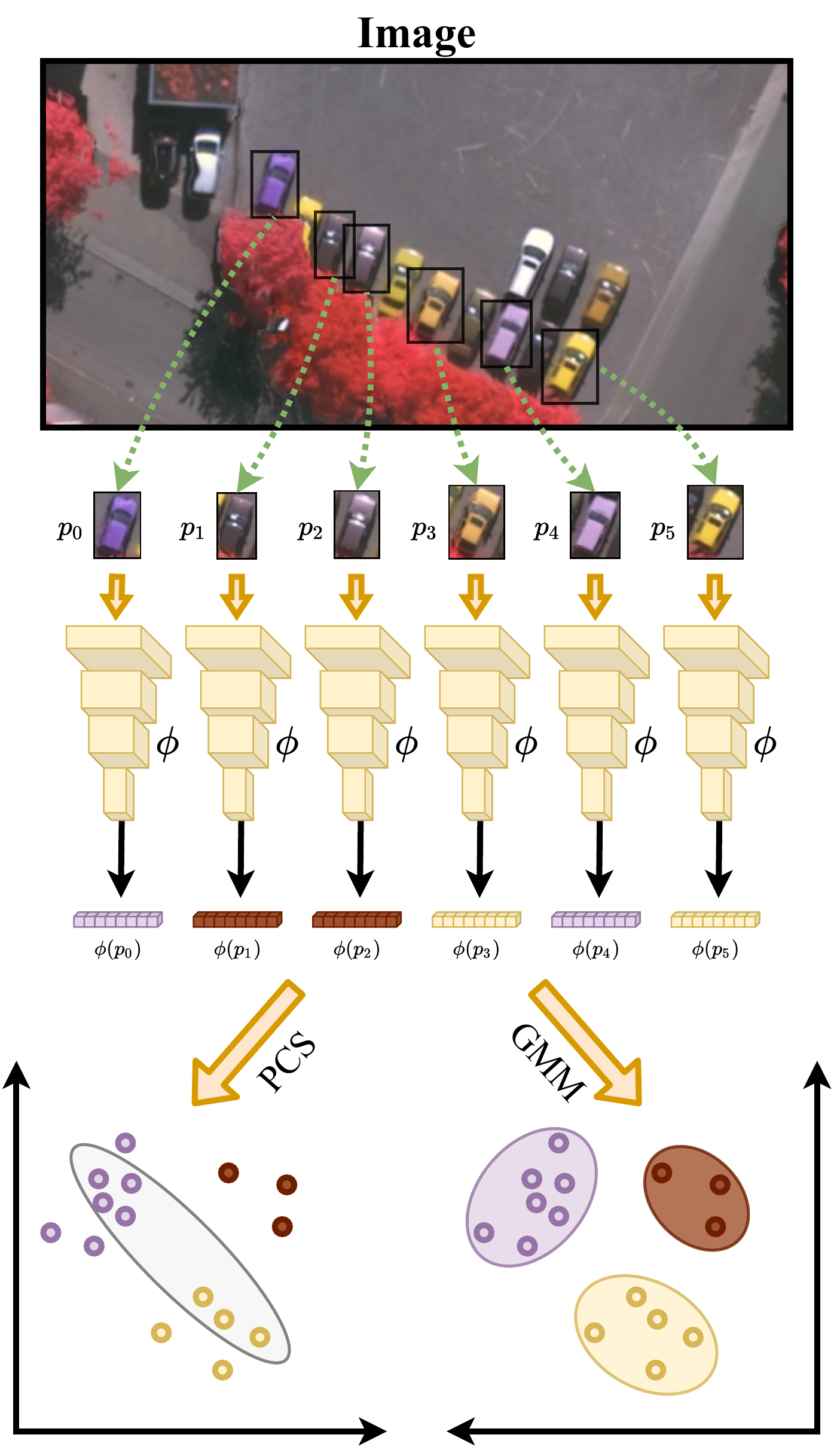}
    \caption{An example showing how different objects can be represented by different distributions. Due to the multimodal representation capability, GMM is better suited for representing real-world data than Principal Component Scoring \cite{oliveira2021fully}.}
    \label{fig:pcs_gmm}
\end{figure}

OpenGMM builds upon the previously proposed OpenPCS \cite{oliveira2021fully} method, OpenPCS uses Principal Component Analysis (PCA) to compress the representation extracted from the backbones and uses the generated representation to do the OOD pixel detection. OpenGMM replaces PCA with Gaussian Mixture Model (GMM) \cite{rasmussen2003gaussian}. GMM's multimodal representation should be better suited for modeling real-world pixels that may not conform to unimodal representations. 

Like OpenPCS, OpenGMM takes into account intermediate feature maps together with the last layer activation map. Combining the activations from the earlier layers with the last layers produces a tensor that fuses low and high semantic level information. The concatenated tensor may have hundreds or thousands of channels, which is known to be redundant \cite{sun2020conditional, huang2017densely}. OpenGMM handles the concatenated tensor size and redundancy of activations by fitting a Gaussian Mixture of Models (GMM) rather than fitting a Principal Component Analysis (PCA) like OpenPCS. The GMM is applied to model each closed set KKC distribution, generating as many models as KKCs. Each GMM model generates a score tensor with the log-likelihood values for all pixels, a final score tensor is generated combining all GMM scores with the closed set prediction. To determine the OOD pixels, the final score is evaluated and pixels below the threshold are set to unknown.

The OpenPCS framework proposed by Oliveira \textit{et al.} \cite{oliveira2021fully} and, as an extention, OpenGMM allow for multiple closed set backbones. We adopt three backbones for OpenGMM and OpenPCS: DenseNet-121 (DN-121) \cite{huang2017densely}, WideResNet-50 (WRN-50) \cite{zagoruyko2016wide} and U-net \cite{ronneberger2015u}. 

Readers should notice that adapting any closed set semantic segmentation network to the OpenGMM and OpenPCS frameworks is relatively quick and simple to implement, without requiring retraining or additional layers to be trained. This is in contrast to other existing OSR/OSS methods \cite{hendrycks2018deep,yoshihashi2019classification,sun2020conditional,nunes2022conditional}. 

The plug-and-play characteristic of methods such as OpenPCS and OpenGMM is a great advantage when considering the problem of adapting the solution to real-world applications and novel domains.

\subsection{Improving Semantic Consistency with Superpixels}
\label{sec:oss_superpixels}

Superpixels are usually employed before or during the segmentation process 
\cite{ji2020parallel,melas2021pixmatch,kang2021picoco,ratajczak2020semantic,zhang2014discriminative }. In general, when employed as post processing, normally they use the input image to generate the SPS and apply it somehow in the output prediction to generate more consistent borders between objects and improve semantic consistency. Following the literature, in the present work we employ superpixels as a post-processing step applied to the OSS scores.

As an output, OSS methods generate a tensor of the same size as the input image containing the scores or reconstruction errors depending on the method used. This final tensor is then processed with the superpixels and all pixels are set to the mean value of the segment from which they belong. Algorithm~\ref{alg:usage} details the usage of the superpixel oversegmentation in the final step just before the open set recognition phase. While our post-processing scheme is agnostic to the choice of SPS algorithm, for this study we evaluated SLIC \cite{achanta2012slic}, Quickshift (QS) \cite{vedaldi2008quick} and Felzenszwalb (FZ) \cite{felzenszwalb2004efficient}. These algorithms were chosen because they have different generation characteristics and different pros and cons, as detailed in Section~\ref{sec:related}. 

\begin{algorithm}
	\caption{The output of the OSS method is post-processed using the superpixels segmentation. All pixels of a given segment assume the mean value of all pixels in that segment.}
	\label{alg:usage}
	\begin{algorithmic}[1]
	    \Require {scores} \Comment{pixelwise array}
	    \Require {segments} \Comment{list of segments}
	    \Procedure {post\_process}{$scores, segments$}
	        \State pred = zeros(scores.size)
	        \For{$segment \in Segments$}
	            \State pred[segment] = mean(scores[segment])
	        \EndFor{}
	        \State \textbf{return} pred
        \EndProcedure{}
	\end{algorithmic} 
\end{algorithm}


The remaining steps of the open set recognition process are kept the same for each OSS method. The superpixel oversegmentations are homogeneous and tend to respect object borders. Applying the superpixels to the score image smooths the segmented areas, aiding the OSS algorithm in avoiding errors within the segmented objects.

Each superpixel algorithm has its own generation characteristics, and the final segmentation reproduces these particularities in the results. We can see in Figure~\ref{fig:merge_superpixels} an illustrative example of two SPS that present different characteristics and may represent better different scenarios. In this example, the SLIC algorithm could better represent textures, while the FZ algorithm could better identify borders, although none of the single SPS could represent the underlying image properly, a great improvement is observed when FuSS is used.

\subsection{\method}
\label{sec:fuss}

All SPS algorithms share the same main goals, namely to generate homogeneous areas and to respect borders between different objects. In other words, their goal is to minimize the intracluster/intrasegment variance while maximizing the intercluster/intersegment variance.

\begin{figure}[!ht]
    \centering
    \includegraphics[width=\columnwidth]{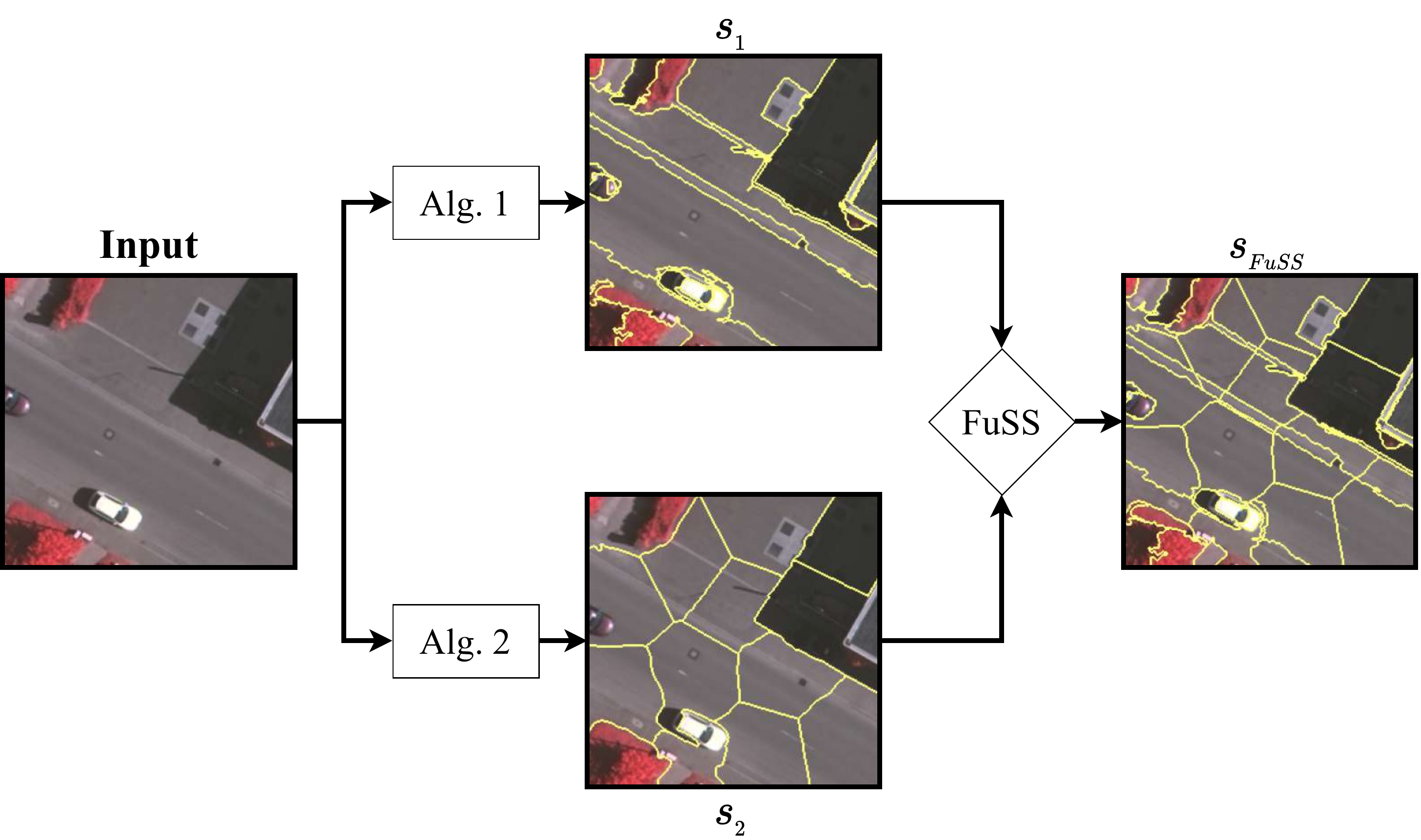}
    \caption{A toy example illustrating the workflow to merge two different superpixel segmentations. First the input image $x$ is processed by 2 different superpixel segmentation algorithms (Alg. 1 and Alg. 2). Then the generated segmentations $s_{_{1}}$ and $s_{_{2}}$ are merged into the final segmentation $s_{_{FuSS}}$ using the merging procedure described in Algorithm~\ref{alg:fuss}.}
    \label{fig:superpixel_workflow}
\end{figure}

\begin{figure*}[!ht]
    \centering
    \includegraphics[width=2.0\columnwidth]{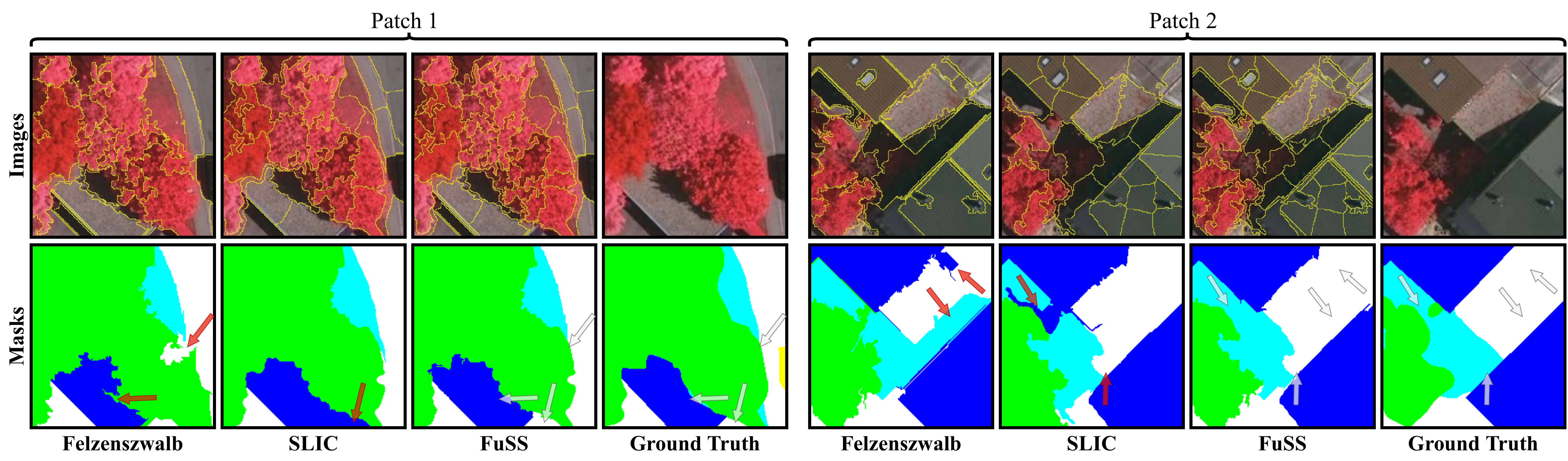}
    \caption{Comparison of the resulting segmentation from two SPS algorithms (Felzenszwalb and SLIC) and our proposed fusion algorithm, \method. The first row shows the input image superimposed with the superpixel segments and the second row depicts the closer class fit of each segment according to the real labels. Red arrows indicate areas where class boundaries failed when using one single SPS algorithm, while gray arrows point to these same regions fixed after applying the \method{} algorithm.}
    \label{fig:merge_superpixels}
\end{figure*}
Different SPS algorithms use distinct operations to produce the final segmentation, and as a result experience distinct failure cases. The idea behind \method{} is to fuse segments from multiple types of superpixel generation algorithms. The motivation for the use of distinct families of SPS methods is to take advantage of the different generation characteristics, enhancing the advantages and mitigating the disadvantages of each method. Figure~\ref{fig:merge_superpixels} shows an example of how each single SPS represents ground truth and compare to \method{} segmentation. To generate a ground truth that perfectly overlaps with all segments we assign the class with highest pixel count to the entire segment for each segment. We can observe through the qualitative result that the \method{} improves the representation of the image in relation to the ground truth. 


The first step of the fusion procedure is to generate unique segments superposing two different segmentations. This procedure is prone to produce some extremely small segments. Thus, to deal with this unwanted side-effect, we use the Malahanobis distance \cite{mahalanobis1936generalized} to fuse the closest neighbor segments until there are no more segments below the specified pixel minimum size. \method{} is detailed in Algorithm~\ref{alg:fuss}.

    

Figure~\ref{fig:superpixel_workflow} illustrates two different superpixels methods being merged. As shown in the figure, the final SPS respects both segmentations' borders, and each segment depict better the underlying region. \method{} is agnostic to the SPS algorithm, being applicable to any set of distinct superpixel algorithms. However, in practice, using more than two algorithms yields exceedingly small segments, motivating our experiments to focus only on pairs of algorithms.

\begin{algorithm}
	\caption{Pseudo-algorithm detailing the auxiliary procedure of joining segmentations used in \method.} 
	\label{alg:fuss}
	\begin{algorithmic}[1]
	    \Require {scores} \Comment{pixelwise array}
	    \Require {segments} \Comment{list of segments}
	    \Procedure {join\_segmentations}{$seg1, seg2$}
	        \State {joint = []}
	        \For{$s1 \in seg1$}
	            \LeftComment {Selecting s2 $\in$ seg2 where s2 $\cap$ s1 $\neq \emptyset$}
	            \For{s2 $\in$ seg2.\Call{overlap\_segments}{s1}}
    	            \State {overlap\_area = $s1 \cap s2$}
    	            \State {joint.\Call{add\_new\_segment}{overlap\_area}}
    	        \EndFor
	        \EndFor
	        \State \textbf{return} joint
        \EndProcedure
        \State{}
        \Procedure {FuSS}{$seg1, seg2$}
	        \State joint = \Call{join\_segmentations}{seg1, seg2}
	        \For {$s \in joint$}
	            \If {$s.size < min\_size$}
	                \State closest = \Call{closest\_neighbor}{s, joint}
	                \State joint = \Call{merge\_segments}{joint, s, closest}
	            \EndIf
            \EndFor
	    \State \textbf{return} joint
	    \EndProcedure
        \end{algorithmic} 
\end{algorithm}
\section{Experimental Setup}
\label{sec:experimental_setup}

We used PyTorch \cite{neurips2019_9015} to implement all neural network models and backbones, with a NVIDIA Titan X with 12GB of memory. All models fit on a single graphics card, filling between 10 GB and 11 GB of memory. SPS algorithms were implemented using the \textit{scikit-learn}\footnotemark\footnotetext{\url{https://scikit-learn.org/}} library. The official implementation for FuSS is publicly available for encouraging reproducibility. 

We tested OpenGMM with 4, 8 and 16 components, resulting in minimal performance differences across this range of values. Hence, all OpenGMM results reported in this section were run with 4 components. For OpenPCS we adopted the 16 components from the original implementation.

\subsection{Datasets and Evaluation Protocol}
\label{sec:datasets}

We employed the Leave One Class Out (LOCO) protocol used by Oliveira \textit{et al.} \cite{oliveira2021fully} in this work to emulate an open set scenario, since the two selected datasets were intended to closed set semantic segmentation. The LOCO protocol splits the known classes and selects one of them to be ignored during training, allowing the evaluation of open set methods on the hidden class. Thus, we only backpropagate the loss of pixels from known classes, ignoring the background, borders, miscellaneous and unknown classes. 

We selected the International Society for Photogrammetry and Remote Sensing (ISPRS) 2D Semantic Labeling\footnotemark\footnotetext{\url{https://www.isprs.org/education/benchmarks/UrbanSemLab/default.aspx}} datasets of Vaihingen and Potsdam for our experiments. Both datasets were previously used in OSS scenarios \cite{da2020towards,oliveira2021fully,nunes2022conditional}. Vaihingen images present a 9cm/pixel spatial resolution, varying from 2000 to 2500 pixels per axis, while Potsdam samples have a 5cm/pixel spatial resolution and $6000\times6000$ pixels each. Both datasets contain 6 KKCs: impervious surfaces, buildings, low vegetation, high vegetation, car and miscellaneous; and 1 known unknown class (KUC): segmentation boundaries between objects. Among KKCs, the miscellaneous class is composed mostly of areas presenting image acquisition noise and objects unimportant to practical remote sensing applications, motivating its removal from our experimental procedure. We used the same bands employed in previous works on OSS \cite{oliveira2021fully,nunes2022conditional}, namely IR-R-G-nDSM.

Both datasets were separated into 3 sets each: training, validating and testing. For Vaihingen the selected patches were: 1, 3, 5, 7, 13, 17, 21, 26, 32 and 37 for training; 11, 15, 28, 30 and 34 for testing; and 23 for validation. For Potsdam the selected patches were: 2\_10, 2\_13, 2\_14, 3\_10, 3\_12, 3\_13, 3\_14, 4\_11, 4\_12, 4\_13, 4\_14, 4\_15, 5\_10, 5\_12, 5\_13, 5\_14, 5\_15, 6\_8, 6\_9, 6\_10, 6\_11, 6\_12, 6\_13, 6\_15, 7\_7, 7\_9, 7\_11, 7\_12 and 7\_13 for training; 2\_11, 2\_12, 4\_10, 5\_11, 6\_7, 7\_8 and 7\_10 for testing; and 3\_11 and 6\_14 for validation.

\subsection{Backbones}
\label{sec:backbones}

All OSS models use a closed set backbone as a starting point to identify the OOD pixels. Relying on previous results \cite{da2020towards,oliveira2021fully}, we used three of the best performing backbones for our experiments: U-Net \cite{ronneberger2015u}, Wide ResNet-50 (WRN50) \cite{zagoruyko2016wide} and DenseNet-121 (DN121) \cite{huang2017densely}. For CoReSeg \cite{nunes2022conditional}, only U-Net was used, since it naturally matches the structural constraint of this method.

\subsection{Metrics}
\label{sec:metrics}

We use the Receiver Operating Characteristic (ROC) curve and the Area Under the ROC (AUROC) improvements as evidence that including superpixels in the inference phase improves OOD recognition. To measure the performace for both KKCs and UUCs together we also employed the Cohen's kappa score  \cite{cohen1960coefficient} ($\kappa$). The use of $\kappa$ allows the assessment of the reliability of the process since it measures the agreement of the methods with the ground truth. Both metrics were used to compare the predictions between the methods with and without superpixel post-processing. 

\subsection{Superpixel Configurations}
\label{sec:superpixel_configs}

We conducted experiments with 70 different superpixel  configurations for the Vaihingen ablation study. From the initial ablation, we selected the 6 best configurations on Vaihingen to run in Potsdam, which is a much larger dataset. The 70 configurations were divided into 2 categories: 
\begin{enumerate}
    \item \textit{single} -- an execution with different parameters of the selected SPS algorithm;
    \item \textit{\method} -- merges 2 different single configurations using the proposed procedure.
\end{enumerate}

For the subset of the 70 superpixel settings that used \method, we tested the use of mean and median when calculating the distance between neighboring superpixel regions. Furthermore, we tested 2 different minimum sizes for the superpixels: 25 and 50 pixels. The implemented SPS algorithms often use the mean to represent each segment or to define the SPS. Thus, we tested the median to find out if it could better represent the image bands.



In the Appendix \ref{app:superpixel_configurations} we list the 11 superpixel generation configurations reported in this work both in Sections~\ref{sec:ablation} and~\ref{sec:results}. The same merge  parameters were used for the \method{} configurations presented in Section~\ref{sec:results}: \textit{mean} as distance to merge superpixels and with a minimum size of 50 pixels.

\begin{table*}[]
    \centering
    \begin{tabular}{c|ccc|ccc|ccc|ccc|ccc|ccc}
        \hline
        & \multicolumn{3}{c|}{\textbf{UUC: imp. surf.}} & \multicolumn{3}{c|}{\textbf{UUC: building}} & \multicolumn{3}{c|}{\textbf{UUC: low veg.}} & \multicolumn{3}{c|}{\textbf{UUC: high veg.}} & \multicolumn{3}{c|}{\textbf{UUC: car}} & \multicolumn{3}{c}{\textbf{Average}} \\ \cline{2-19} 
        \multirow{-2}{*}{\textbf{Type}} & \multicolumn{1}{c}{\textbf{min}} & \multicolumn{1}{c}{\textbf{avg.}} & \multicolumn{1}{c|}{\textbf{max}} & \multicolumn{1}{c}{\textbf{min}} & \multicolumn{1}{c}{\textbf{avg.}} & \multicolumn{1}{c|}{\textbf{max}} & \multicolumn{1}{c}{\textbf{min}} & \multicolumn{1}{c}{\textbf{avg.}} & \multicolumn{1}{c|}{\textbf{max}} & \multicolumn{1}{c}{\textbf{min}} & \multicolumn{1}{c}{\textbf{avg.}} & \multicolumn{1}{c|}{\textbf{max}} & \multicolumn{1}{c}{\textbf{min}} & \multicolumn{1}{c}{\textbf{avg.}} & \multicolumn{1}{c|}{\textbf{max}} & \multicolumn{1}{c}{\textbf{min}} & \multicolumn{1}{c}{\textbf{avg.}} & \multicolumn{1}{c}{\textbf{max}} \\ \hline
        \textbf{single} & 
        {.860} & {.898} & \textbf{.912} & 
        {.934} & {.956} & \textbf{.961} & 
        {.702} & {.730} & \textbf{.740} & 
        {.824} & {.873} & .886 & 
        {.709} & {.872} & \textbf{.918} & 
        {.830} & {.866} & .878 \\
        \textbf{\method} & 
        {\textbf{.903}} & {\textbf{.907}} & .911 & 
        \textbf{.957} & \textbf{.960} & \textbf{.961} & 
        \textbf{.729} & \textbf{.735} & \textbf{.740} & 
        \textbf{.878} & \textbf{.884} & \textbf{.887} & 
        \textbf{.813} & \textbf{.884} & .913 & 
        \textbf{.860} & \textbf{.874} & .\textbf{888} \\ \hline
    \end{tabular}%
    \caption{The table presents the results aggregated by type of superpixel generation. Two types of generation were used, ``single'' and ``\method''. ``Single'' stands for a generation using one superpixel method alone, while ``\method'' stands for the procedure presented in Section~\ref{sec:fuss}. The columns min, avg. and max stand for the minimum, average and maximum AUROC value among all superpixel configurations for each type of generation. The last block is the is the average value among all tested UUCs.
    }
    \label{tab:ablation_superpixels}
\end{table*}
\section{Ablation Study}
\label{sec:ablation}

\begin{table*}[]
\centering
\begin{tabular}{rrrrrlcc}
\hline
\multicolumn{5}{c}{\textbf{UUCs}} & \multicolumn{1}{c}{\multirow{2}{*}{\textbf{Average}}} & \multirow{2}{*}{\textbf{Superpixel All}} & \multirow{2}{*}{\textbf{Value Used}} \\ \cline{1-5}
\multicolumn{1}{l}{\textbf{Imp. Surf.}} & \multicolumn{1}{l}{\textbf{Building}} & \multicolumn{1}{l}{\textbf{Low Veg.}} & \multicolumn{1}{l}{\textbf{High Veg.}} & \multicolumn{1}{l}{\textbf{Car}} & \multicolumn{1}{c}{} &  &  \\ \hline
.903 & .958 & \textbf{.737} & \textbf{.881} & \textbf{.897} & $\mathbf{.8752 \pm .083}$ & Yes & Mean \\ 
\textbf{.906} & \textbf{.959} & .730 & .880 & .865 & $.8676 \pm .085$ & Yes & Median \\ 
.903 & .958 & \textbf{.737} & \textbf{.881} & \textbf{.897} & $\mathbf{.8752 \pm .083}$ & No & Mean \\ 
\textbf{.906} & \textbf{.959} & .730 & .880 & .865 & $.8677 \pm .085$ & No & Median \\ \hline
\end{tabular}
\caption{The table shows the AUROC results aggregated by the use in post-processing. The columns ''superpixel All`` indicates if the SPS was applied to the entire predictions or only to the scores and the "Value Used" column shows which value is used to represent the entire superpixel.}
\label{tab:ablation_post_process}
\end{table*}

\begin{figure}[!t]
    \centering
    \includegraphics[width=0.8\columnwidth]{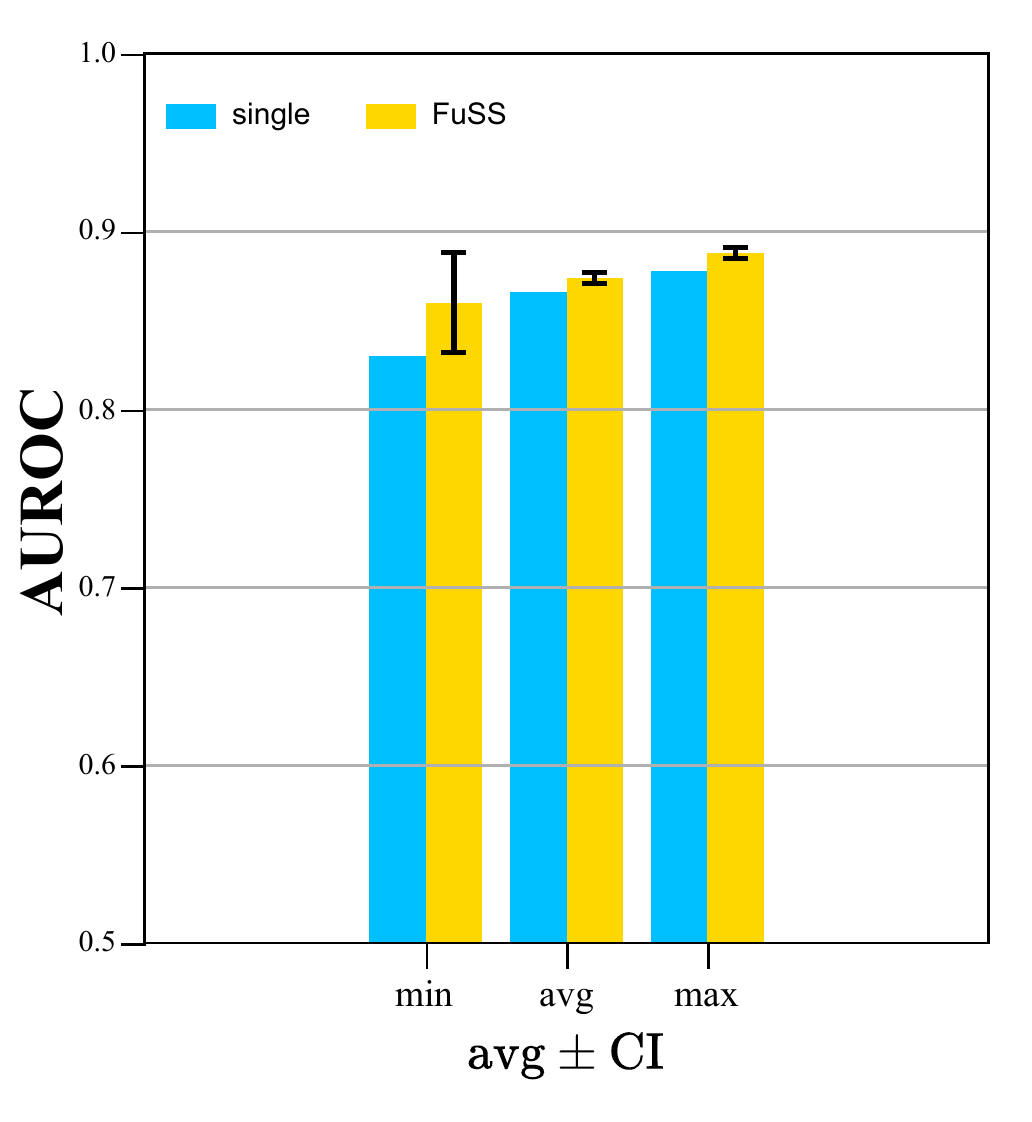}
    \caption{Comparison of minimum, average and maximum AUROC for all superpixel configurations generated for the ablation study on CoReSeg \cite{nunes2022conditional} with the Vaihingen dataset. The barplot shows single superpixel algorithms (yellow) and \method{} (blue) for the average across all UUCs in the LOCO protocol. Confidence Intervals (CIs) according to a paired two-tailed t-Student test with $p \le 0.05$ across the 5 classes are shown as error bars, highlighting the statistical significance of employing \method{} instead of single SPS algorithms. The lower y-axis is trimmed to 0.5 for better visualization of the CIs.}
    \label{fig:superpixels_results_comparison}
\end{figure}

Besides the 70 different configurations mentioned in Section~\ref{sec:experimental_setup}, we applied the superpixels in 4 distinct manners, adding up to 280 different configurations. We executed the ablation study for the Vaihingen dataset using CoReSeg \cite{nunes2022conditional} which has a fixed architecture. OpenPCS and OpenGMM can easily switch backbones and already have 3 backbones tested each, to run the ablation tests for OpenGMM and OpenPCS would be prohibitive as 280 tests would need to be run for each configuration.

We split the following section into two sets of experiments: 1) SPS generation (Section~\ref{sec:ablation_generation}), and 2) Post-Processing for OSS (Section~\ref{sec:ablation_use}).

\subsection{Superpixel Generation}
\label{sec:ablation_generation}

\begin{figure}[!t]
    \centering
    \includegraphics[width=0.8\columnwidth]{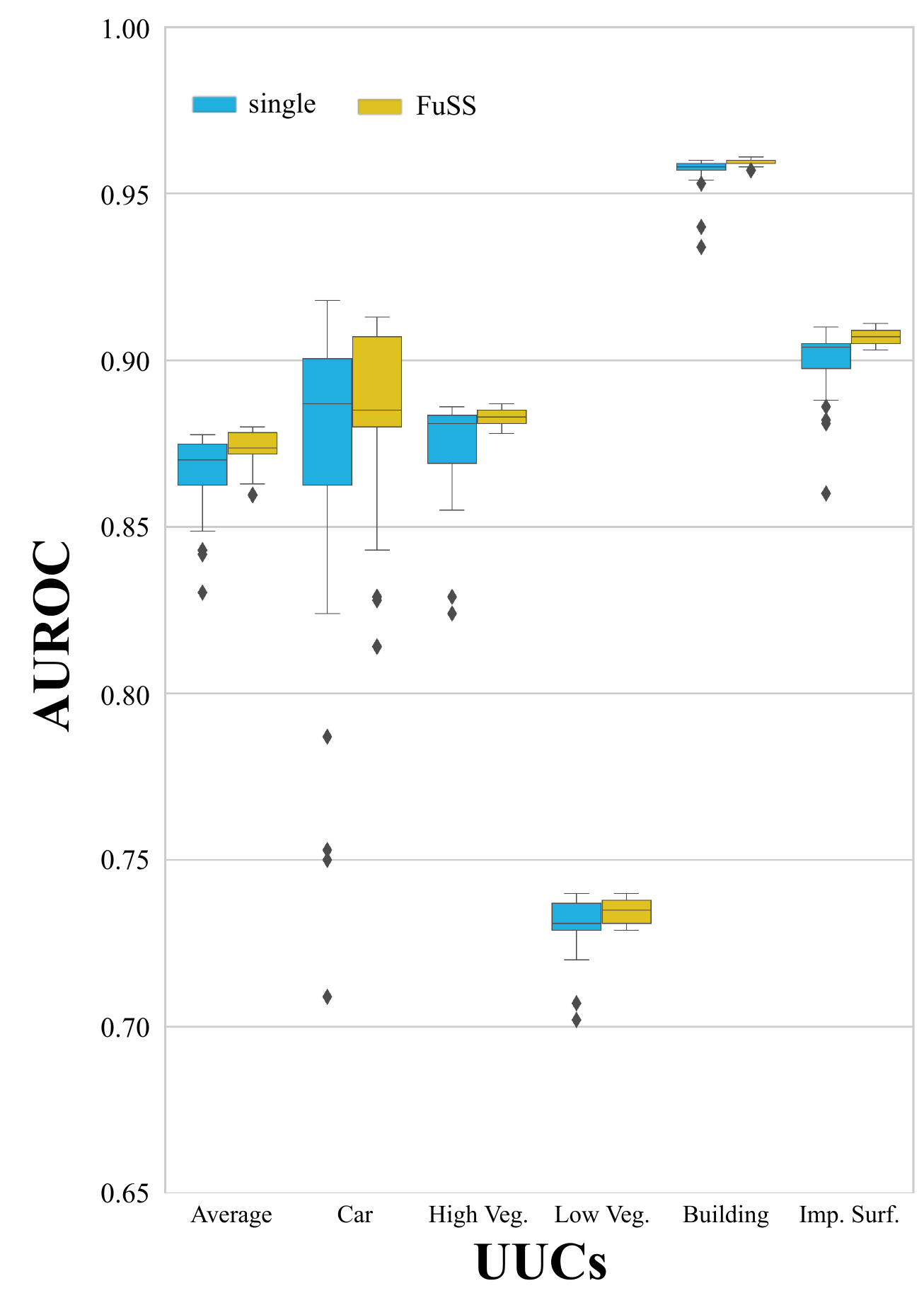}
    \caption{Boxplot with the AUROC results for all configurations used for the ablation study for CoReSeg in the Vaihingen dataset. The boxplot shows both single superpixel algorithms (blue) and \method{} (yellow) for individual UUCs and for the average.}
    \label{fig:ablation_boxplot}
\end{figure}

Table~\ref{tab:ablation_superpixels} and Figure~\ref{fig:ablation_boxplot} present the results aggregated by generation strategy. The results shows that \method{} achieved considerably stabler results in average and regardless of the UUC as well. A key point to use superpixels is to select good parameters for the superpixel generation. \method{} combines different SPSs and made the selection of parameters less relevant to the process. Thus, despite the parameter selection, the combination of 2 different oversegmentations generates a more reliable result. 
Figure~\ref{fig:superpixels_results_comparison} presents the results for the minimum, average and maximum results for both strategies of generation. The error bars show the confidence intervals according to a paired two-tailed t-Student test with $p \le 0.05$ across all 5 tested scenarios.

The use of mean or median to represent the entire superpixel when calculating the distances had no influence in the results, as the superpixel resultant segmentations were similar or even equal in all cases. The same behavior was perceived when considering the minimum size for the segments. This behavior implies that the objects may be represented with either 25 or 50 pixels as the value of minimum size. We can see at Table~\ref{tab:ablation} in Appendix~\ref{app:superpixel_configurations} a list with the best 20 results among all executed tests for this ablation study.

Regarding the \method{} procedure, the ablation study performed in Vaihingen shows that the method achieve stabler results. The worst and the best results are closer when compared to using a single superpixel method as shown in Figure~\ref{fig:ablation_boxplot}. When using any superpixel method, a key part of the process is the selection of the parameters. In this regard, using \method{} relieves this burden, since it combines two different methods and generates segments that fits the input image better. Based on our observations, a conjecture to be studied further is that the more different the input SPSs are, the better the final merged superpixel segmentation is.

\subsection{Post-Processing for OSS}
\label{sec:ablation_use}

Regarding the use of the superpixel segmentation to improve the results of the OSS methods, we tested 4 different ways:
\begin{itemize}
    \item applying to entire closed set segmentation and to the scores/reconstruction errors;
    \item applying only to the scores/reconstruction errors;
    \item use the mean to represent the score/reconstruction error for the entire superpixel;
    \item use the median to represent the score/reconstruction error for the entire superpixel.
\end{itemize}

Table~\ref{tab:ablation_post_process} presents the results aggregated by the ways superpixels were used in the OSS post-process. The best results were achieve using the mean value of the score/reconstruction error to represent the superpixel. Applying the superpixel only to the scores/reconstruction errors or along the closed set prediction resulted in the same performance, implying that the variations are equivalent.



\section{Results and Discussion}
\label{sec:results}

\begin{table*}[]
    \centering
    \begin{tabular}{cccccccc|cccccc}
        \hline
         &  & \multicolumn{6}{c|}{\textbf{Vaihingen}} & \multicolumn{6}{c}{\textbf{Potsdam}} \\ \cline{3-14} 
         &  & \multicolumn{5}{c}{\textbf{UUCs}} &  & \multicolumn{5}{c}{\textbf{UUCs}} &  \\ \cline{3-7} \cline{9-13}
        \multirow{-3}{*}{\textbf{Backbone}} & \multirow{-3}{*}{\textbf{OSS method}} & \textbf{0} & \textbf{1} & \textbf{2} & \textbf{3} & \textbf{4} & \multirow{-2}{*}{\textbf{Average}} & \textbf{0} & \textbf{1} & \textbf{2} & \textbf{3} & \textbf{4} & \multirow{-2}{*}{\textbf{Average}} \\ \hline
        DN121 & OpenGMM & \textbf{.872} & \textbf{.936} & \textbf{.646} & .688 & \textbf{.687} & $\mathbf{.7658 \pm .129}$ & .829 & \textbf{.907} & \textbf{.642} & .553 & .866 & $\mathbf{.7594 \pm .154}$ \\ 
        DN121 & OpenPCA & .860 & .925 & .643 & \textbf{.708} & .650 & $.7572 \pm .128$ & \textbf{.841} & .901 & .546 & \textbf{.569} & \textbf{.882} & $.7478 \pm .175$ \\ \hline
        U-Net & CoReSeg & .884 & \textbf{.934} & \textbf{.710} & \textbf{.867} & \textbf{.854} & $\mathbf{.8499 \pm .084}$ & .824 & \textbf{.901} & \textbf{.698} & \textbf{.631} & \textbf{.768} & $\mathbf{.7644 \pm .106}$ \\ 
        U-Net & OpenGMM & \textbf{.908} & .787 & .522 & .846 & .601 & $.7328 \pm .165$ & \textbf{.870} & .856 & .371 & .569 & .858 & $.7048 \pm .226$ \\ 
        U-Net & OpenPCA & .899 & .799 & .508 & .843 & .538 & $.7174 \pm .181$ & .875 & .859 & .423 & .531 & .835 & $.7046 \pm .212$ \\ \hline
        WRN50 & OpenGMM & \textbf{.885} & \textbf{.911} & .611 & .648 & .619 & $\mathbf{.7348 \pm .150}$ & .851 & .870 & .403 & .453 & .824 & $.6802 \pm .231$ \\ 
        WRN50 & OpenPCA & .854 & .873 & \textbf{.628} & \textbf{.658} & \textbf{.622} & $.7270 \pm .126$ & \textbf{.853} & \textbf{.880} & \textbf{.424} & \textbf{.462} & \textbf{.840} & $\mathbf{.6918 \pm .228}$ \\ \hline
    \end{tabular}
    \caption{For each combination of backbone and OSS method the table shows the classwise AUROC and average (Avg.) AUROC in Vaihingen and Potsdam. Numbered columns stand for: 1 – Impervious Surfaces, 2 – Building, 3 – Low Vegetation, 4 – High Vegetation and 5 – Car}
    \label{tab:results_gmm}
\end{table*}

\subsection{OpenGMM Comparisons}
\label{sec:opengmm_results}

Table~\ref{tab:results_gmm} presents comparisons between OpenGMM, OpenPCS and CoReSeg with the same closed set backbones. In Table~\ref{tab:results_gmm} for each backbone the cells in bold shows the baseline results before any post-processing. The post-processing results will be discussed in Section~\ref{sec:superpixels_post_processing}.

For the Vaihingen dataset we can observe that in all tested cases OpenGMM outperformed OpenPCS with the same backbone, while Potsdam yielded mixed results, with OpenGMM and OpenPCS presenting overall similar performances and even surpassing OpenPCS in 4 out of 6 direct comparisons.
We attribute the improvement shown by OpenGMM over OpenPCS to its multimodal representation capability for modeling real-world data, 
regardless of the number of KKCs. The worse results in Potsdam are attributed mainly to OpenGMM's poorer performances on two UUCs: Low Vegetation and High Vegetation. The instability of OSS algorithms in these two particular classes is known from previous works, possibly due to the large semantic intra-class variabilities.

It is worth mentioning that our experimental procedure is not exactly the same used by Oliveira \textit{et al.} \cite{oliveira2021fully}, therefore, all experiments were re-executed to ensure comparability and the results presented here are different from the previous work.


\subsection{Post-Processing OSS with Superpixels}
\label{sec:superpixels_post_processing}

\begin{figure*}[!t]
    \centering
    \includegraphics[width=\textwidth]{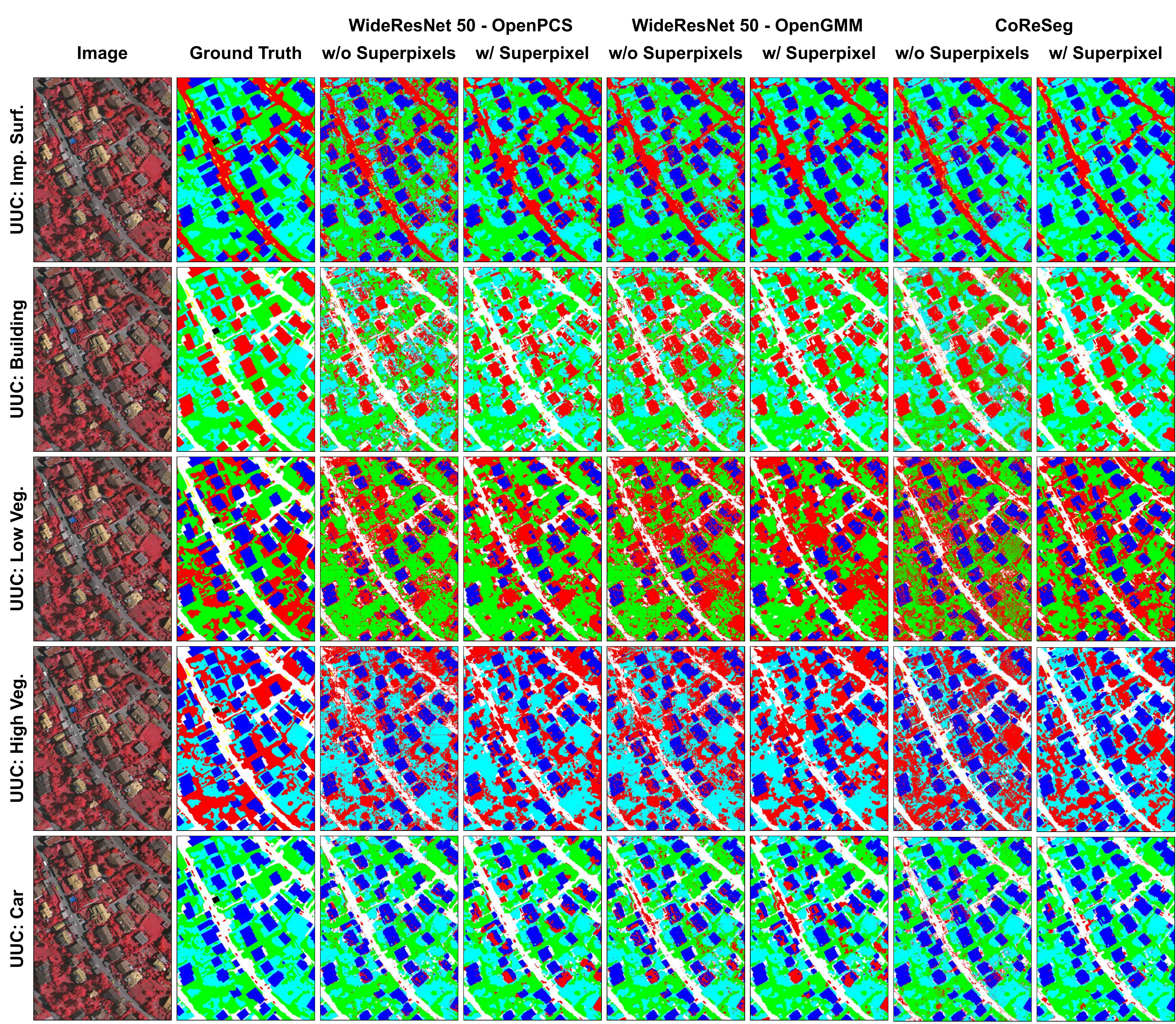}
    \caption{Qualitative results for an image from the Vaihingen dataset under different settings of UUCs and OSS methods. The use of the proposed superpixel post-processing method generates clearer segmentations, avoiding the usual mislabeling of unkown pixels. }
    \label{fig:comparative}
\end{figure*}

\begin{figure*}[!t]
    \centering
    \includegraphics[width=0.85\textwidth]{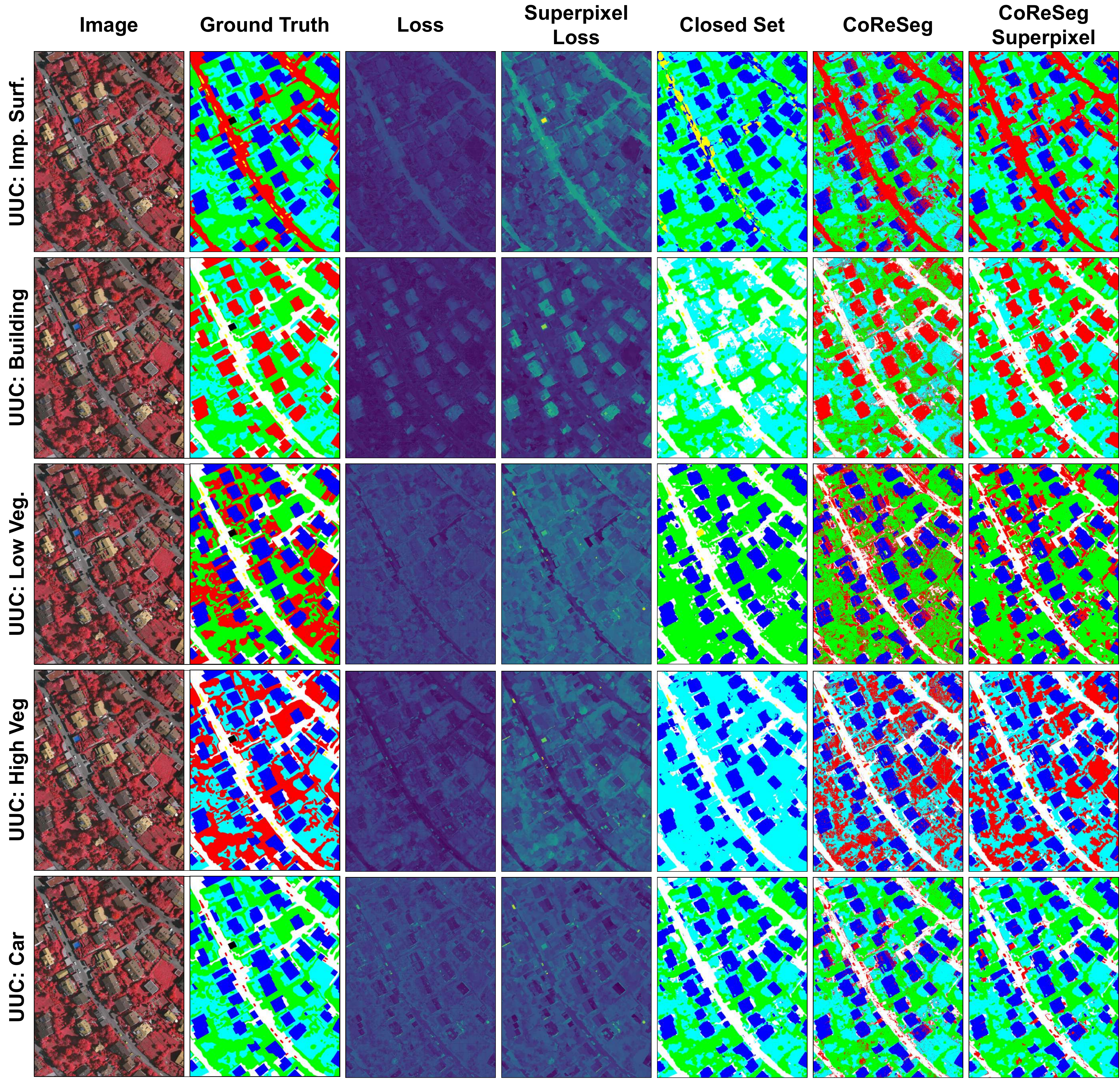}
    \caption{OSS predictions obtained using CoReSeg for an image of the Vaihingen dataset with all tested UUCs. The last column shows qualitatively the improvement achieved using the proposed superpixel post-processing method. We can also observe the impact of the method in the Reconstruction Loss.}
    \label{fig:vaihingen_images_coreseg}
\end{figure*}

Superpixel post-processing improved AUROC in all tested scenarios. The improvement and the better semantic consistency can be clearly observed in 
Figures~\ref{fig:comparative} and~\ref{fig:vaihingen_images_coreseg}.

Tables~\ref{tab:potsdam_results} and~\ref{tab:vaihingen_results} show, in terms of AUROC, the improvement observed in the quantitative analysis. We only observe a worsening in the AUROC average values for the cases in which the original closed set result of the OSS method was already poor for a certain UUC. Table~\ref{tab:vaihingen_results} shows this for both OpenGMM and OpenPCS for UUCs Low Vegetation and Car with U-Net as backbone.

From the results, we can state that if the OSS results presents little semantic consistency, the post-processing is not able to consistently improve the results. In fact, the experiments show that the post-processing can worsen the results in such cases where prediction performance is already rather poor. However, when the OSS model with no superpixel post-processing yields consistent segmentations with inconsistent predictions mainly in object borders and/or salt-and-pepper artifacts, superpixel post-processing was quite effective. In other words, assuming that the superpixels are representative, homogeneous and respect the edges of the image, a better base OSS result allows the superpixel post-processing to correct mistakes inside the superpixels, as these mistakes are usually the minority of pixels. For instance, considering CoReSeg with U-Net as backbone and Car as the UUC, the AUROC improved from 0.854 to 0.913 for Vaihingen, and from 0.768 to 0.816 for Potsdam. Another clear improvement can be seen when analysing the average AUROC for the 5 UUCs, which improved from 0.850 to 0.880 in Vaihingen and from 0.764 to 0.797 in Potsdam.


Tables~\ref{tab:results_vaihingen_openpcs_opengmm_kappa} and~\ref{tab:results_vaihingen_coreseg_kappa} show the comparison of the Cohen's Kappa scores for different thresholds on the tested OSS methods. 
$\kappa$ improved for all thresholds in the results presented in Table~\ref{tab:results_vaihingen_coreseg_kappa}. Metrics followed by $\dagger$ show statistically significant improvements with the use of superpixels, which was achieved for 8 out of 11 thresholds in CoReSeg. Considering the OpenPCS and OpenGMM performances presented in Table~\ref{tab:results_vaihingen_openpcs_opengmm_kappa}, the results of thresholds lower than 0.5 were improved, but they were not statistically significant. As for thresholds between 0.5 and 0.9, the $\kappa$ improvement was statistically significant. The proposed post-processing was able not only to improve the AUROC threshold independent score and the qualitative results, but the results for Cohen's kappa score reassures that the post-processing improves the reliability of the predictions.

\begin{table*}[]
    \centering
    \begin{tabular}{cccccccccc}
        \hline
        \multicolumn{1}{c}{} & \multicolumn{1}{c}{} & \multicolumn{1}{c}{} & \multicolumn{5}{c}{\textbf{UUCs}} & \multicolumn{1}{c}{} & \multicolumn{1}{c}{} \\ \cline{4-8}
        \multicolumn{1}{c}{\multirow{-2}{*}{\textbf{Backbone}}} & \multicolumn{1}{c}{\multirow{-2}{*}{\textbf{OSS Method}}} & \multicolumn{1}{c}{\multirow{-2}{*}{\textbf{Config.}}} & \multicolumn{1}{c}{\textbf{Imp. Surf.}} & \multicolumn{1}{c}{\textbf{Buildings}} & \multicolumn{1}{c}{\textbf{Low Veg.}} & \multicolumn{1}{c}{\textbf{High Veg.}} & \multicolumn{1}{c}{\textbf{Car}} & \multicolumn{1}{c}{\multirow{-2}{*}{\textbf{avg. AUROC}}} & \multicolumn{1}{c}{\multirow{-2}{*}{\textbf{avg. px/seg}}} \\ \hline
        DN121 & OpenGMM & - & .872 & .936 & .646 & .688 & .687 & $.7658 \pm .129$ & - \\
        DN121 & OpenGMM & fuss01 & \textbf{.891$ \dagger$} & .946 & \textbf{.652$ \dagger$} & \textbf{.698} & .689 & $.7752 \pm .133$ & 750 \\
        DN121 & OpenGMM & single02 & .880 & \textbf{.954$ \dagger$} & .650 & .696 & \textbf{.701$ \dagger$} & $ \mathbf{.7762 \pm .133 \dagger}$ & 491 \\ \hline 
        DN121 & OpenPCS & - & .860 & .925 & .643 & .708 & .650 & $.7572 \pm .128$ & - \\
        DN121 & OpenPCS & fuss02 & \textbf{.878} & .948 & \textbf{.651} & \textbf{.720} & .636 & $.7694 \pm .135$ & 322 \\
        DN121 & OpenPCS & single02 & .876 & \textbf{.951} & .649 & .718 & \textbf{.663} & $.7714 \pm .135$ & 491 \\ \hline \hline
        U-Net & CoReSeg & - & .884 & .934 & .710 & .867 & .854 & $.8499 \pm .084$ & - \\
        U-Net & CoReSeg & single03 & .906 & .958 & .737 & .884 & .903 & $.8776 \pm .083$ & 476 \\
        U-Net & CoReSeg & fuss03 & .906 & \textbf{.959} & \textbf{.739} & \textbf{.886} & \textbf{.910} & $ \mathbf{.8800 \pm .083} $ & 434 \\ \hline
        U-Net & OpenGMM & - & .908 & .787 & .522 & .846 & .601 $ \dagger$ & $.7328 \pm .165$ & - \\
        U-Net & OpenGMM & fuss01 & \textbf{.930}$ \dagger$ & .805 & .512 & .846 & .597 & $.7380 \pm .176$ & 750 \\
        U-Net & OpenGMM & single01 & .929 & .802 & .519$ \dagger$ & .848 & .599 & $.7394 \pm .173 \dagger$ & 322 \\ \hline
        U-Net & OpenPCS & - & .899 & .799 & .508 & .843 & .538 & $.7174 \pm .181$ & - \\
        U-Net & OpenPCS & single04 & .925 & .814 & .501 & .848 & .527 & $.7230 \pm .195$ & 322 \\
        U-Net & OpenPCS & fuss02 & .925 & .813 & .502 & .852 & .526 & $.7236 \pm .196$ & 492 \\ \hline \hline
        WRN50 & OpenGMM & - & .885 & .911 & .611 & .648 & .619 & $.7348 \pm .150$ & - \\
        WRN50 & OpenGMM & single01 & .904 & \textbf{.936}$ \dagger$ & .622 & .665 & .618 & $.7490 \pm .165$ & 491 \\
        WRN50 & OpenGMM & fuss04 & \textbf{.918}$ \dagger$ & \textbf{.936}$ \dagger$ & .630 & .667 & \textbf{.641} $ \dagger$ & $ \mathbf{.7584 \pm .155}$ $ \dagger$ & 698  \\ \hline
        WRN50 & OpenPCS & - & .854 & .873 & .628 & .658 & .622 & $.7270 \pm .126$ & - \\
        WRN50 & OpenPCS & fuss04 & .896 & .915 & .628 & \textbf{.681} & .571 & $.7378 \pm .158$ & 698 \\
        WRN50 & OpenPCS & single02 & .890 & .912 & \textbf{.648} & \textbf{.681} & .600 & $.7462 \pm .144$ & 491 \\ \hline \hline
    \end{tabular}%
    \caption{Results for the Vaihingen dataset ordered by the average AUROC for each pair Backbone/OSS Method. Each configuration of the column Superpixel Config. is better detailed in the Appendix \ref{app:superpixel_configurations}. The column avg. AUROC shows the average AUROC between the UUCs; the column avg. px/seg shows the average size of the segments generated by the Superpixel Config. The $ \dagger$ symbol marks when results for OpenGMM is better than OpenPCS with the same Backbone-OSS method.}
    \label{tab:vaihingen_results}
\end{table*}

\begin{table*}[]
    \centering
    \begin{tabular}{cccccccccc}
        \hline
        \multicolumn{1}{c}{} & \multicolumn{1}{c}{} & \multicolumn{1}{c}{} & \multicolumn{5}{c}{\textbf{UUCs}} & \multicolumn{1}{c}{} & \multicolumn{1}{c}{} \\ \cline{4-8}
        \multicolumn{1}{c}{\multirow{-2}{*}{\textbf{Backbone}}} & \multicolumn{1}{c}{\multirow{-2}{*}{\textbf{OSS Method}}} & \multicolumn{1}{c}{\multirow{-2}{*}{\textbf{Config.}}} & \multicolumn{1}{c}{\textbf{Imp. Surf.}} & \multicolumn{1}{c}{\textbf{Buildings}} & \multicolumn{1}{c}{\textbf{Low Veg.}} & \multicolumn{1}{c}{\textbf{High Veg.}} & \multicolumn{1}{c}{\textbf{Car}} & \multicolumn{1}{c}{\multirow{-2}{*}{\textbf{avg. AUROC}}} & \multicolumn{1}{c}{\multirow{-2}{*}{\textbf{avg. px/seg}}} \\ \hline
        DN121 & OpenGMM & - & .829 & .907 & .642 & .553 & .866 & $.7594 \pm .154$ & - \\
        DN121 & OpenGMM & single03 & .852 & .914 & \textbf{.642}$ \dagger$ & .559 & .871 & $.7676 \pm .154$ & 509 \\
        DN121 & OpenGMM & fuss01 & .848 & \textbf{.916}$ \dagger$ & \textbf{.642}$ \dagger$ & .567 & .887 & $ \mathbf{.7720 \pm .157} \dagger$ & 1006 \\ \hline
        DN121 & OpenPCS & - & .841 & .901 & .546 & .569 & .882 & $.7478 \pm .175$ & - \\
        DN121 & OpenPCS & single01 & \textbf{.866} & .913 & .569 & .571 & .888 & $.7614 \pm .177$ & 876 \\
        DN121 & OpenPCS & fuss01 & .863 & .913 & .558 & \textbf{.582} & \textbf{.904} & $.7640 \pm .178$ & 1006 \\ \hline \hline
        U-Net & CoReSeg & - & .824 & .901 & .698 & .631 & .768 & $.7644 \pm .106$ & - \\
        U-Net & CoReSeg & single05 & .860 & \textbf{.913} & \textbf{.738} & .654 & .813 & $.7956 \pm .102$ & 1447 \\
        U-Net & CoReSeg & fuss01 & .866 & .911 & .733 & \textbf{.659} & .817 & $ \mathbf{.7972 \pm .102}$ & 1006 \\ \hline
        U-Net & OpenGMM & - & .870 & .856 & .371 & .569 & .858 & $.7048 \pm .226$ & - \\
        U-Net & OpenGMM & single03 & .898 & .880 & .357 & .571 & .876 & $.7164 \pm .243$ & 509 \\
        U-Net & OpenGMM & fuss01 & .895 & .878 & .361 & .578$ \dagger$ & \textbf{.897}$ \dagger$ & $.7218 \pm .243$ & 1006 \\ \hline
        U-Net & OpenPCS & - & .875 & .859 & .423 & .531 & .835 & $.7046 \pm .212$ & - \\
        U-Net & OpenPCS & single03 & .901 & .893 & .415 & .524 & .864 & $.7194 \pm .232$ & 509 \\
        U-Net & OpenPCS & fuss04 & \textbf{.902} & .892 & .416 & .526 & .874 & $.7220 \pm .233$ & 876 \\ \hline \hline
        WRN50 & OpenGMM & - & .851 & .870 & .403 & .453 & .824 & $.6802 \pm .231$ & - \\
        WRN50 & OpenGMM & single03 & .884 & .890 & .418 & .455 & .835 & $.6964 \pm .239$ & 509 \\
        WRN50 & OpenGMM & fuss01 & .880 & .890 & .413 & .457 & .853 & $.6986 \pm .242$ & 1006 \\ \hline
        WRN50 & OpenPCS & - & .853 & .880 & \textbf{.424} & \textbf{.462} & .840 & $.6918 \pm .228$ & - \\
        WRN50 & OpenPCS & single03 & \textbf{.888} & \textbf{.904} & .423 & .457 & .856 & $.7058 \pm .243$ & 509 \\
        WRN50 & OpenPCS & fuss01 & .884 & \textbf{.904} & .419 & .457 & \textbf{.873} & $ \mathbf{.7074 \pm .247}$ & 1006 \\ \hline \hline
    \end{tabular}%
    \caption{Results for the Potsdam dataset ordered by the average AUROC for each pair Backbone/OSS Method. Each configuration of the column Superpixel Config. is better detailed in the Appendix  \ref{app:superpixel_configurations}. The column avg. AUROC shows the average AUROC between the UUCs; the column avg. px/seg shows the average size of the segments generated by the Superpixel Config. The $\dagger$ symbol marks when results for OpenGMM is better than OpenPCS with the same Backbone.}
    \label{tab:potsdam_results}
\end{table*}

\begin{table*}[]
    \centering
    \begin{tabular}{ccccccccccccc}
        \hline\hline
        \multicolumn{1}{c}{\multirow{2}{*}{\textbf{Method}}} &
        \multicolumn{1}{c}{\multirow{2}{*}{\textbf{UUC}}} & \multicolumn{1}{c}{\multirow{2}{*}{\textbf{\method}}} & \multicolumn{10}{c}{\textbf{Thresholds}} \\ \cline{4-13} 
        \multicolumn{1}{c}{} & \multicolumn{1}{c}{} & \multicolumn{1}{c}{} & \multicolumn{1}{c}{\textbf{0.00}} & \multicolumn{1}{c}{\textbf{0.10}} & \multicolumn{1}{c}{\textbf{0.20}} & \multicolumn{1}{c}{\textbf{0.30}} & \multicolumn{1}{c}{\textbf{0.40}} & \multicolumn{1}{c}{\textbf{0.50}} & \multicolumn{1}{c}{\textbf{0.60}} & \multicolumn{1}{c}{\textbf{0.70}} & \multicolumn{1}{c}{\textbf{0.80}} & \multicolumn{1}{c}{\textbf{0.90}} \\ \hline
        OpenGMM & Imp. Surf. & No & \textbf{0.500} & 0.500 & 0.510 & 0.568 & 0.616 & 0.644 & 0.665 & 0.670 & 0.642 & 0.541\\ 
        OpenGMM & Imp. Surf. & Yes & \textbf{0.500} & \textbf{0.530} & \textbf{0.560} & \textbf{0.591} & \textbf{0.621} & \textbf{0.649} & \textbf{0.676} & \textbf{0.693} & \textbf{0.687} & \textbf{0.617} \\ \hline
        OpenGMM & Buildings & No & \textbf{0.461} & 0.461 & 0.481 & 0.532 & 0.581 & 0.608 & 0.631 & \textbf{0.649} & 0.654 & 0.612 \\ 
        OpenGMM & Buildings & Yes & \textbf{0.461} & \textbf{0.493} & \textbf{0.524} & \textbf{0.555} & \textbf{0.586} & \textbf{0.615} & \textbf{0.643} & 0.664 & \textbf{0.678} & \textbf{0.652} \\ \hline
        OpenGMM & Low Veg. & No & \textbf{0.606} & \textbf{0.575} & \textbf{0.567} & \textbf{0.551} & \textbf{0.525} & 0.490 & 0.436 & 0.365 & 0.278 & 0.177 \\
        OpenGMM & Low Veg. & Yes & \textbf{0.606} & 0.564 & 0.558 & 0.541 & \textbf{0.525} & \textbf{0.496} & \textbf{0.462} & \textbf{0.405} & \textbf{0.325} & \textbf{0.218} \\ \hline
        OpenGMM & High Veg. & No & \textbf{0.518} & 0.509 & 0.523 & 0.534 & 0.536 & 0.520 & 0.484 & 0.418 & 0.304 & 0.166 \\ 
        OpenGMM & High Veg. & Yes & \textbf{0.518} & \textbf{0.512} & \textbf{0.526} & \textbf{0.537} & \textbf{0.539} & \textbf{0.525} & \textbf{0.498} & \textbf{0.443} & \textbf{0.368} & \textbf{0.217} \\ \hline
        OpenGMM & Car & No & \textbf{0.775} & \textbf{0.747} & \textbf{0.712} & \textbf{0.672} & \textbf{0.607} & \textbf{0.515} & 0.393 & 0.281 & 0.175 & 0.073 \\ 
        OpenGMM & Car & Yes & \textbf{0.775} & 0.735 & 0.705 & 0.655 & 0.573 & 0.500 & \textbf{0.414} & \textbf{0.324} & \textbf{0.223} & \textbf{0.131} \\ \hline
        OpenGMM & Average & No & \textbf{0.572} & 0.558 & 0.559 & 0.571 & \textbf{0.573} & 0.555 & 0.522 & 0.477 & 0.411 & 0.314 \\ 
        OpenGMM & Average & Yes & \textbf{0.572} & \textbf{0.567} & \textbf{0.575} & \textbf{0.576} & 0.569 & \textbf{0.557} & \textbf{0.539}$\dagger$ & \textbf{0.506}$\dagger$ & \textbf{0.456}$\dagger$ & \textbf{0.367}$\dagger$ \\ \hline\hline
        OpenPCS & Imp. Surf. & No & \textbf{0.500} & \textbf{0.525} & \textbf{0.552} & 0.578 & 0.602 & 0.622 & 0.635 & 0.634 & 0.607 & 0.519 \\ 
        OpenPCS & Imp. Surf. & Yes & \textbf{0.500} & 0.522 & 0.550 & \textbf{0.583} & \textbf{0.613} & \textbf{0.638} & \textbf{0.656} & \textbf{0.661} & \textbf{0.648} & \textbf{0.575} \\ \hline
        OpenPCS & Buildings & No & \textbf{0.461} & 0.466 & 0.520 & 0.543 & 0.566 & 0.586 & 0.599 & 0.605 & 0.598 & 0.551 \\ 
        OpenPCS & Buildings & Yes & \textbf{0.461} & \textbf{0.472} & \textbf{0.524} & \textbf{0.551} & \textbf{0.579} & \textbf{0.607} & \textbf{0.627} & \textbf{0.632} & \textbf{0.640} & \textbf{0.603} \\ \hline
        OpenPCS & Low Veg. & No & \textbf{0.606} & \textbf{0.606} & \textbf{0.604} & \textbf{0.589} & 0.547 & 0.478 & 0.406 & 0.332 & 0.293 & 0.245 \\ 
        OpenPCS & Low Veg. & Yes & \textbf{0.606} & 0.590 & 0.592 & 0.585 & \textbf{0.562} & \textbf{0.532} & \textbf{0.463} & \textbf{0.367} & \textbf{0.310} & \textbf{0.254}  \\ \hline
        OpenPCS & High Veg. & No & \textbf{0.518} & 0.511 & 0.529 & 0.541 & 0.542 & 0.527 & 0.489 & 0.424 & 0.319 & 0.168 \\ 
        OpenPCS & High Veg. & Yes & \textbf{0.518} & \textbf{0.517} & \textbf{0.541} & \textbf{0.554} & \textbf{0.556} & \textbf{0.540} & \textbf{0.505} & \textbf{0.456} & \textbf{0.379} & \textbf{0.213} \\ \hline
        OpenPCS & Car & No & \textbf{0.775} & \textbf{0.771} & \textbf{0.763} & \textbf{0.732} & \textbf{0.623} & \textbf{0.468} & \textbf{0.369} & \textbf{0.293} & \textbf{0.198} & \textbf{0.109}  \\ 
        OpenPCS & Car & Yes & \textbf{0.775} & 0.748 & 0.723 & 0.674 & 0.573 & 0.443 & 0.332 & 0.237 & 0.154 & 0.086  \\ \hline
        OpenPCS & Average & No & \textbf{0.572} & \textbf{0.576} & \textbf{0.594} & \textbf{0.597} & \textbf{0.576} & 0.536 & 0.500 & 0.458 & 0.403 & 0.319  \\ 
        OpenPCS & Average & Yes & \textbf{0.572} & 0.570 & 0.586 & 0.590 & \textbf{0.576} & \textbf{0.552}$\dagger$ & \textbf{0.517}$\dagger$ & \textbf{0.471}$\dagger$ & \textbf{0.426}$\dagger$ & \textbf{0.346}$\dagger$  \\ \hline\hline
    \end{tabular}%
    \caption{Kappa scores with thresholds varying from 0.0 to 1.0 for the Vaihingen dataset with OpenGMM and OpenPCS as the OSS method and WRN50 as backbone. The third column represents wether the superpixel post-processing was applied. 
    The last two rows for each combination of backbone and OSS method shows the average kappa value across all UUCs. For the average rows, the $\dagger$ symbol means that the results has statistical significance, according to a paired two-tailed t-Student test with $p \le 0.05$ across the 5 classes.}
    \label{tab:results_vaihingen_openpcs_opengmm_kappa}
\end{table*}

\begin{table*}[]
    \centering
    \begin{tabular}{ccccccccccccc}
        \hline\hline
        \multicolumn{1}{c}{\multirow{2}{*}{\textbf{UUC}}} & \multicolumn{1}{c}{\multirow{2}{*}{\textbf{\method{}}}} & \multicolumn{11}{c}{\textbf{Thresholds}} \\ \cline{3-13} 
        \multicolumn{1}{c}{} & \multicolumn{1}{c}{} & \multicolumn{1}{c}{\textbf{0.90}} & \multicolumn{1}{c}{\textbf{0.91}} & \multicolumn{1}{c}{\textbf{0.92}} & \multicolumn{1}{c}{\textbf{0.93}} & \multicolumn{1}{c}{\textbf{0.94}} & \multicolumn{1}{c}{\textbf{0.95}} & \multicolumn{1}{c}{\textbf{0.96}} & \multicolumn{1}{c}{\textbf{0.97}} &  \multicolumn{1}{c}{\textbf{0.98}} &
        \multicolumn{1}{c}{\textbf{0.99}} &\multicolumn{1}{c}{\textbf{1.00}} \\ \hline
        Imp. Surfaces & No & 0.678 & 0.683 & 0.687 & 0.692 & 0.696 & 0.699 & 0.701 & 0.698& 0.685 & 0.564 & \textbf{0.540} \\
        Imp. Surfaces & Yes & \textbf{0.703} & \textbf{0.705} & \textbf{0.708} & \textbf{0.710} & \textbf{0.710} & \textbf{0.710} & \textbf{0.712} & \textbf{0.711}  & \textbf{0.700} & \textbf{0.549} & \textbf{0.540} \\ \hline
        Buildings & No & 0.692 & 0.698 & 0.703 & 0.707 & 0.710 & 0.709 & 0.705 & 0.695 & 0.674 & 0.623 & \textbf{0.504} \\
        Buildings & Yes & \textbf{0.744} & \textbf{0.746} & \textbf{0.748} & \textbf{0.749} & \textbf{0.747} & \textbf{0.742} & \textbf{0.733} & \textbf{0.717} & \textbf{0.689} & \textbf{0.626} & \textbf{0.504} \\\hline
        Low Vegetation & No & 0.587 & 0.589 & 0.591 & 0.592 & 0.594 & 0.595 & 0.597 & 0.600 & 0.605 & 0.611 & \textbf{0.621} \\
        Low Vegetation & Yes & \textbf{0.600} & \textbf{0.601} & \textbf{0.604} & \textbf{0.603} & \textbf{0.605} & \textbf{0.606} & \textbf{0.607} & \textbf{0.609} & \textbf{0.611} & \textbf{0.615} & \textbf{0.621} \\\hline
        High Vegetation & No & 0.656 & 0.657 & 0.658 & 0.658 & 0.656 & 0.653 & 0.648 & 0.638 & \textbf{0.619} & \textbf{0.555} & \textbf{0.555} \\
        High Vegetation & Yes & \textbf{0.683} & \textbf{0.684} & \textbf{0.683} & \textbf{0.683} & \textbf{0.681} & \textbf{0.676} & \textbf{0.665} & \textbf{0.645} & 0.609 & 0.549 & \textbf{0.555} \\\hline
        Car & No & 0.677 & 0.684 & 0.692 & 0.701 & 0.710 & 0.720 & 0.732 & 0.745 & 0.757 & 0.769 & 0.778 \\
        Car & Yes & \textbf{0.713} & \textbf{0.719} & \textbf{0.726} & \textbf{0.733} & \textbf{0.741} & \textbf{0.749} & \textbf{0.758} & \textbf{0.764} & \textbf{0.769} & \textbf{0.773} & \textbf{0.779} \\\hline
        Average & No & \multicolumn{1}{c}{0.658} & \multicolumn{1}{c}{0.662} & \multicolumn{1}{c}{0.666} & \multicolumn{1}{c}{0.670} & \multicolumn{1}{c}{0.673} & \multicolumn{1}{c}{0.675} & \multicolumn{1}{c}{0.677} & \multicolumn{1}{c}{0.675} & \multicolumn{1}{c}{0.644} & \multicolumn{1}{c}{0.649} & \multicolumn{1}{c}{\textbf{0.600}} \\
        Average & Yes & \textbf{0.689} $ \dagger$ & \textbf{0.691}$ \dagger$ & \textbf{0.694}$ \dagger$ & \textbf{0.695}$ \dagger$ & \textbf{0.697}$ \dagger$ & \textbf{0.697}$ \dagger$ & \textbf{0.695}$ \dagger$ & \textbf{0.689}$ \dagger$ & \textbf{0.645} & \textbf{0.652} & \textbf{0.600} \\ \hline\hline
    \end{tabular}%
    \caption{Kappa scores with thresholds varying from 0.9 to 1.0 for the Vaihingen dataset with CoReSeg as the OSS method. The second column represents wether the superpixel post-processing was applied. For the average rows, the $\dagger$ symbol means that the results has statistical significance, according to a paired two-tailed t-Student test with $p \le 0.05$ across the 5 classes.}
    \label{tab:results_vaihingen_coreseg_kappa}
\end{table*}
\section{Conclusion}
\label{sec:conclusion}

In this paper, we presented two different approaches to improve the semantic consistency for the results of OSS. The first one is an extension of OpenPCS \cite{oliveira2021fully} by replacing the unimodal Principal Component model by a multimodal Gaussian Mixture of Models. The second one is a superpixel post-processing pipeline capable of benefiting OSS predictions based on single SPS algorithms or a mixture of them in a novel SPS fusion method named \method. 

OpenGMM improved OpenPCS results for most of the UUCs and most of tested backbones in terms of AUROC average value. 
Furthermore, the superpixel post-processing method achieved state-of-the-art results for both Vaihingen and Potsdam datasets when applied to CoReSeg, improving the average AUROC in 0.030 
for Vaihingen and 0.033 
for Potsdam. 
The improvement produced by the post-processing was statistically significant in the majority of tested cases. 

Our novel \method{} fusion scheme uses Malahanobis distance to merge neighboring segments below the established minimum size limit. 
\method{} produced more stable and reliable SPS compared to the use of a single superpixel method. 
Furthermore, in most cases the best results were achieved using the final segmentation generated by \method. 
\method was also proven to relieve the burden of parameter selection for superpixel generation. 

Our proposed SPS post-processing method improved the results in all tested scenarios and for all OSS methods and backbones. This study shows that methods aiming to improve semantic consistency can benefit from a superpixel post-processing procedure, which helps in delimiting objects and borders.

Future works include: evaluating other contour recognition post-processing methods; training a deep neural network that generates superpixels simultaneously with the semantic segmentation task \cite{yang2020superpixel}; testing different modern state-of-the-art segmentation backbones \cite{wang2020deep} with OpenPCS and OpenGMM; employing attention techniques for the improvement of the OSS; and using conditional random fields as the post processing step at the end of the networks.

\clearpage
\appendices

\section{Superpixel Configurations}
\label{app:superpixel_configurations}

Table~\ref{tab:ablation} is complimentary to Section\ref{sec:ablation} and shows the top 20 results for Vaihingen dataset and CoReSeg \cite{nunes2022conditional} among the 280 tests executed. The first line shows the baseline without post-processing, and the subsequent lines are sorted by Average AUROC. The second column stands for the value used to represent the superpixel for post-processing, and the third column shows the minimum pixel count of each superpixel used by FuSS.

Below the list of all superpixels configurations used to run all tests in this work presented in Sections~\ref{sec:results} and \ref{sec:ablation} and in Table~\ref{tab:ablation}. FZ stands for the Felzenszwalb \textit{et al.} \cite{felzenszwalb2004efficient} algorithm, QS stands for the Quickshift \cite{vedaldi2008quick} algorithm, and SLIC stands for the method with the same name proposed by Achanta \textit{et al.} \cite{achanta2012slic}:
\begin{enumerate}
    \item \textit{single01}: FZ (scale: 100, sigma: 0.5, min\_size: 50)
    \item \textit{single02}: FZ (scale: 200, sigma: 0.5, min\_size: 50)
    \item \textit{single03}: SLIC (n\_segments: n\_pixels $\div$ 350, compactness: 5, $\sigma$: 1)
    \item \textit{single04}: FZ (scale: 50, sigma: 0.5, min\_size: 50)
    \item \textit{single05}: FZ (scale: 100, sigma: 0.5, min\_size: 100)
    \item \textit{fuss01}: 
    \begin{itemize}
        \item SLIC (n\_segments: n\_pixels $\div$ 2000, compactness: 5, $\sigma$: 1)
        \item FZ (scale: 200, $\sigma$: 0.7, min\_size: 200)
    \end{itemize}  
    \item \textit{fuss02}: 
    \begin{itemize}
        \item SLIC (n\_segments: n\_pixels $\div$ 1500, compactness: 5, $\sigma$: 1)
        \item FZ (scale: 100, $\sigma$: 0.7, min\_size: 150)
    \end{itemize}  
    \item \textit{fuss03}: 
    \begin{itemize}
        \item SLIC (n\_segments: n\_pixels $\div$ 1000, compactness: 5, $\sigma$: 1)
        \item FZ (scale: 100, $\sigma$: 0.7, min\_size: 150)
    \end{itemize}  
    \item \textit{fuss04}: 
    \begin{itemize}
        \item FZ (scale: 200, $\sigma$: 0.7, min\_size: 200)
        \item QS (kernel\_size: 5, max\_dist: 50, ratio: 0.5)
    \end{itemize}  
    \item \textit{fuss05}: 
    \begin{itemize}
        \item FZ (scale: 200, $\sigma$: 0.7, min\_size: 200)
        \item QS (kernel\_size: 4, max\_dist: 50, ratio: 0.5)
    \end{itemize}  
    \item \textit{fuss06}: 
    \begin{itemize}
        \item FZ (scale: 200, $\sigma$: 0.7, min\_size: 200)
        \item QS (kernel\_size: 3, max\_dist: 50, ratio: 0.5)
    \end{itemize}  
\end{enumerate}

\begin{table*}[!t]
    \centering
    \begin{tabular}{cccccccccc}
        \hline
        \multicolumn{1}{c}{\multirow{2}{*}{\textbf{Config}}} &
        \multicolumn{1}{c}{\multirow{2}{*}{\textbf{Value Used}}} &
        \multicolumn{1}{c}{\multirow{2}{*}{\textbf{min. size}}} & \multicolumn{5}{c}{\textbf{UUCs}} & \multicolumn{1}{c}{\multirow{2}{*}{\textbf{avg. AUROC}}} 
        \\ \cline{4-8}
        \multicolumn{1}{c}{} & \multicolumn{1}{c}{}& \multicolumn{1}{c}{} & \multicolumn{1}{c}{\textbf{Imp. Surf.}} & \multicolumn{1}{c}{\textbf{Buildings}} & \multicolumn{1}{c}{\textbf{Low Veg.}} & \multicolumn{1}{c}{\textbf{High Veg.}} & \multicolumn{1}{c}{\textbf{Car}} & \multicolumn{1}{c}{} 
        \\ \hline
        \multicolumn{1}{c}{-} & \multicolumn{1}{c}{-} & \multicolumn{1}{c}{-} & .884 & .934 & .710 & .867 & .854 & $.8499 \pm .084$ & 
        \\
        fuss03 & mean & 50 & \textbf{906} & .959 & .739 & \textbf{.886} & .910 & $\mathbf{.8800 \pm .083}$ 
        \\
        fuss03 & median & 50 & \textbf{.906} & .959 & .739 & \textbf{.886} & .910 & $\mathbf{.8800 \pm .083}$ 
        \\
        fuss03 & mean & 25 & \textbf{.906} & .959 & .739 & \textbf{.886} & .910 & $.8797 \pm .083$ 
        \\
        fuss03 & median & 25 & .905 & .959 & .739 & \textbf{.886} & .910 & $.8797 \pm .083$ 
        \\
        fuss02 & median & 50 & \textbf{.906} & .959 & \textbf{.740} & .885 & .908 & $.8796 \pm .083$ 
        \\
        fuss01 & mean & 50 & .904 & \textbf{.960} & .739 & .882 & \textbf{.913} & $.8796 \pm .084$ 
        \\
        fuss02 & mean & 50 & \textbf{.906} & .959 & \textbf{.740} & .885 & .908 & $.8796 \pm .083$ 
        \\
        fuss01 & median & 50 & .904 & \textbf{.960} & .739 & .882 & \textbf{.913} & $.8796 \pm .084$ 
        \\
        fuss01 & median & 25 & .904 & \textbf{.960} & .738 & .882 & .912 & $.8794 \pm .084$ 
        \\
        fuss01 & mean & 25 & .904 & \textbf{.960} & .738 & .882 & .912 & $.8794 \pm .084$ 
        \\
        fuss02 & median & 25 & \textbf{.906} & .959 & .739 & .885 & .907 & $.8794 \pm .083$ 
        \\
        fuss02 & mean & 25 & \textbf{.906} & .959 & .739 & .885 & .907 & $.8793 \pm .083$ 
        \\
        fuss05 & median & 50 & .905 & \textbf{.960} & .739 & .881 & .910 & $.8789 \pm .083$ 
        \\
        fuss06 & mean & 50 & .905 & .959 & .738 & .883 & .909 & $.8788 \pm .083$ 
        \\
        fuss05 & mean & 50 & .905 & \textbf{.960} & .739 & .881 & .909 & $.8788 \pm .083$ 
        \\
        fuss06 & median & 50 & .905 & .959 & .738 & .883 & .908 & $.8787 \pm .083$ 
        \\
        fuss04 & mean & 50 & .904 & \textbf{.960} & .739 & .880 & .911 & $.8786 \pm .083$ 
        \\
        fuss04 & median & 50 & .904 & \textbf{.960} & .739 & .880 & .911 & $.8786 \pm .083$ 
        \\
        fuss05 & median & 25 & .904 & .959 & .739 & .881 & .909 & $.8785 \pm .083$ 
        \\
        fuss05 & mean & 25 & .904 & .959 & .739 & .881 & .909 & $.8785 \pm .083$ 
        \\
        fuss06 & mean & 25 & .905 & .959 & .738 & .883 & .907 & $.8783 \pm .083$ 
        \\
        fuss04 & median & 25 & .903 & \textbf{.960} & .739 & .880 & .910 & $.8783 \pm .083$ 
    \end{tabular}%
    \caption{Top 20 average AUROC results for all five UUCs tested scenarios achieved with different superpixel post-processing methods for OSS, obtained with CoReSeg \cite{nunes2022conditional} on the Vaihingen dataset. The first line shows the baseline result without post-processing. The first column presents the superpixel segmentation config used, the second shows which value was chosen to represent the superpixel to calculate the distance between 2 superpixels, the third the minimum pixel count for each superpixel, and the last column the the average AUROC for the 5 tested scenarios.}
    \label{tab:ablation}
\end{table*}

\clearpage
\bibliography{refs.bib}{}
\bibliographystyle{IEEEtran}
\begin{IEEEbiography}[{\includegraphics[width=1in,height=1.25in,clip,keepaspectratio]{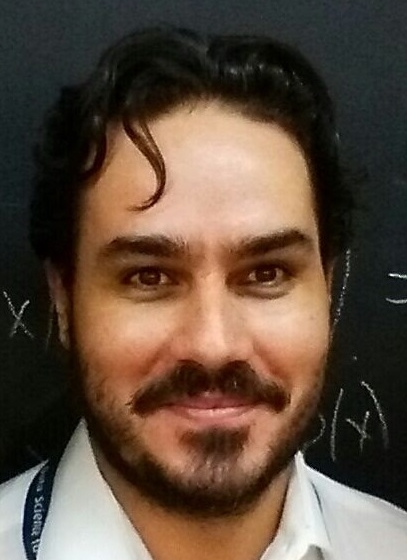}}]{Ian Monteiro Nunes} currently applying for the PhD degree at the Department of Informatics at Pontifical Catholic University of Rio de Janeiro (PUC-Rio), Rio de Janeiro - Brasil, where he also completed the MSc. degree in 2008 and the BSc. degree in Computer Engeneering in 2003. Research areas include Machine Learning, Deep Learning, Domain Adaptation, Computer Vision, Remote Sensing, Clustering and Open Set Recognition.
\end{IEEEbiography}

\begin{IEEEbiography}[{\includegraphics[width=1in,height=1.25in,clip,keepaspectratio]{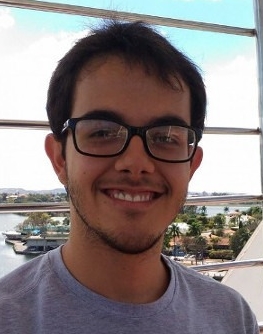}}]{Matheus Pereira} currently pursuing the PhD degree at the Department of Computer Science at Universidade Federal de Minas Gerais (UFMG), Belo Horizonte - Brazil, where he also completed the MSc. degree in 2019. Bsc. (2017) degree in Computer Science by Universidade Federal de Lavras (UFLA). Research areas include Machine Learning, Computer Vision, Remote Sensing, Long-Tailed Recognition and Semantic Segmentation.
\end{IEEEbiography}

\begin{IEEEbiography}[{\includegraphics[width=1in,height=1.25in,clip,keepaspectratio]{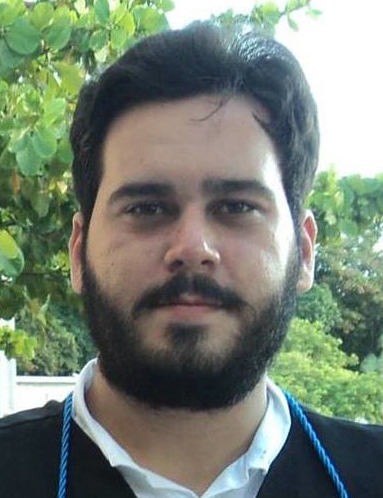}}]{Hugo Oliveira} currently a Post-Doctoral fellow at the Institute of Mathematics and Statistics at Universidade de São Paulo (IME/USP). BSc. (2014) and MSc. (2016) degrees in Computer Science by Universidade Federal da Paraíba (UFPB). PhD in Computer Science (2020) by Universidade Federal de Minas Gerais (UFMG), Belo Horizonte, Brazil. Research areas include Machine Learning, Deep Learning, Domain Adaptation, Deep Generative Models, Few-Shot Learning, Meta-Learning, Metric-Learning, Self-Supervised Learning and Open Set Recognition.
\end{IEEEbiography}

\begin{IEEEbiography}[{\includegraphics[width=1in,height=1.25in,clip,keepaspectratio]{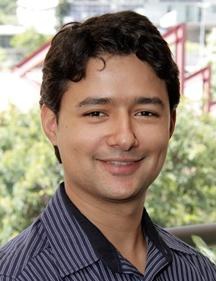}}]{Jefersson Alex dos Santos} has got a Ph.D in Computer Science from Université de Cergy-Pontoise (France) and from University of Campinas (Brazil) in 2013. Currently, he is an Adjunct Professor in the Department of Computer Science at the Universidade Federal de Minas Gerais, Brazil.
He has a Research Productivity scholarship from the Brazilian Research Council (CNPq) since 2016.
Jefersson has published several articles in journals with high impact factor and selective editorial policy. He has also published more than thirty articles in important conferences of remote sensing, image processing and computer vision areas.
Jefersson has experience in coordinating research with Brazilian funding agencies and R\&D projects with companies in those topics. Jefersson is founder and coordinator of the Laboratory of Pattern Recognition and Earth Observation (PATREO - www.patreo.dcc.ufmg.br), one of Brazil's pioneer groups focused on the development of computer vision and machine learning for remote sensing applications.
\end{IEEEbiography}

\begin{IEEEbiography}[{\includegraphics[width=1in,height=1.25in,clip,keepaspectratio]{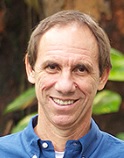}}]{Marcus Poggi} is an Associate Professor at PUC-Rio and his main area of expertise is Computer Science with emphasis on Theory of Computation and Optimization and Automated Reasoning. In these areas he conducts research in Industrial Automation, Decision Support Systems, Operational Research, Design andAnalysis of Algorithm at the Galgos laboratory. He has an undergraduate degree in Electrical Engineering from Pontifícia Universidade Católica do Rio de Janeiro (1983), a master’s degree in Electrical Engineering also from PUC-Rio (1988) and a PhD in Applied Mathematics from Ecole Polytechnique de Montreal (1993).
\end{IEEEbiography}
\EOD
\end{document}